\documentclass{interact} 

\pdfoutput=1

\usepackage{epstopdf}
\usepackage[caption=false]{subfig}

\usepackage{wrapfig}

\usepackage[natbibapa,nodoi]{apacite}
\newcommand{\new}[1]{{#1}} 
\newcommand{\edit}[1]{{#1}}
\newcommand{\upd}[1]{{#1}} 
\newcommand{\rev}[1]{{\color{black} #1}} 
\usepackage{amsmath}
\usepackage{subfig}
\usepackage{booktabs}
\usepackage{hyperref}
\usepackage{balance}
\usepackage{xcolor}
\usepackage{graphicx}
\usepackage[font={small}]{caption}
\usepackage{multirow}

\begin{document}


\title{\edit{Moving fast and slow}:\\ Analysis of representations and post-processing in speech-driven automatic gesture generation
}



\author{
\name{Taras Kucherenko\textsuperscript{a}\thanks{Corresponding author: Taras Kucherenko. Email: \href{mailto:tarask@kth.se}{tarask@kth.se}}, Dai Hasegawa\textsuperscript{b}, Naoshi Kaneko\textsuperscript{c}, Gustav Eje Henter\textsuperscript{a} and Hedvig Kjellstr{\"o}m\textsuperscript{a}}
\affil{\textsuperscript{a}KTH Royal Institute of Technology, Stockholm, Sweden; \textsuperscript{b}Hokkai Gakuen University, Sapporo, Japan; \textsuperscript{c}Aoyama Gakuin University, Sagamihara, Japan}
}

\maketitle

\begin{abstract}
This paper presents a novel framework for speech-driven
gesture production, 
applicable to \edit{virtual agents to enhance} human-computer interaction. Specifically, we extend recent deep-learning-based,
data-driven methods for speech-driven gesture generation by
incorporating representation learning.
Our model takes speech as input and produces gestures as output, in the form of a sequence of 3D coordinates.

We provide an analysis of different representations for the input (speech) and the output (motion) of the network by both objective and subjective evaluations. We also analyse the importance of smoothing of the produced motion.

Our results indicated that the proposed method improved on our baseline in terms of objective measures. For example, it better captured the motion dynamics and better matched the motion-speed distribution.
Moreover, we performed user studies on two different datasets. The studies confirmed that our proposed method is perceived as more natural than the baseline, although the difference in the studies was eliminated by appropriate post-processing: hip-centering and smoothing.
We conclude that it is important to take both \rev{motion representation} and post-processing into account when designing an automatic gesture-production method.



\begin{keywords}
Gesture generation; representation learning; neural network; deep learning; virtual agents; non-verbal behavior
\end{keywords}

\end{abstract}

\section{Introduction}
\label{sec:intro}
Conversational agents in the form of virtual agents or social robots are becoming increasingly widespread and may soon enter our day-to-day lives. Humans use non-verbal behaviors to signal their intent, emotions and attitudes in human-human interactions \citep{knapp2013nonverbal,matsumoto2013nonverbal}. In the same vein, it has been shown that people read and interpret robots' non-verbal cues, e.g., gestures and gaze, similarly to how people read non-verbal cues from each other \citep{breazeal2005effects}. Equipping robots with such non-verbal behaviors have shown to positively affect people's perception of the robot \citep{salem2013err,rifinski2020human}.
\upd{It is important that conversational agents produce appropriate communicative non-verbal behaviors, such that a viewer can easily interpret the communicative intent/function of the behaviors. }


An important part of non-verbal communication is gesticulation: gestures made with hands, arms, head pose and body pose convey a large share of non-verbal content \citep{mcneill1992hand}. To facilitate natural human-agent interaction, it is hence important to enable robots and embodied virtual agents to accompany their speech with gestures in the way people do. We, therefore, explore hand-gesture generation for virtual agents.

\edit{Traditionally, hand gestures have been generated using rule-based methods} \citep{cassell2001beat, ng2010synchronized, huang2012robot}. These methods are rather rigid as they can only generate gestures that are incorporated in the rules. Writing down rules for all plausible gestures found in human interaction is impossible. Consequently, it is difficult to fully capture the richness of human gesticulation in rule-based systems. \edit{Recently, data-driven approaches such as \citet{hasegawa2018evaluation, ginosar2019learning, sadoughi2017speech} have offered }a solution toward eliminating this bottleneck of the rule-based approach. \edit{We continue this line of research.}
In this paper, we present \new{a data-driven method that learns to generate human motions (specifically beat gestures) from a dataset of human actions.} In particular, we use speech data as input, since it is correlated with hand gestures \citep{mcneill1992hand} and has the same temporal character.



To predict gestures from speech, we apply Deep Neural Networks (DNNs), which have been widely used in human skeleton modeling \citep{martinez2017human, butepage2017deep, kucherenko2018neural}.
We further apply representation learning on top of conventional speech-input, gesture-output DNNs.
Representation learning is a branch of unsupervised learning 
aiming to learn a better representation of the data.
Typically, representation learning is applied to make a subsequent learning task easier.
Inspired by previous successful applications to 
learning human motion dynamics, for example in prediction \citep{butepage2017deep} 
and motion synthesis \citep{habibie2017recurrent}, this paper \edit{extends a previous approach for neural-network-based speech-to-gesture mapping by \citet{hasegawa2018evaluation} by applying representation learning to the motion sequence}.

The contributions of this paper are three-fold: 
\begin{enumerate}
    \item We propose a novel speech-driven gesture-production method: \rev{Aud2Repr2Pose}.
    \item We evaluate the importance of representation both for the output motion (by comparing different motion representations) 
    and for the input speech (by comparing different speech feature representations). 
    \item We also evaluate the effect of post-processing of the produced gestures.
\end{enumerate}

The evaluation used two methodologies: First, we evaluated speech-feature selection and choice of architecture for the motion-representation model, \new{comparing generated and ground-truth signals in terms of body-joint positions and motion statistics.}
Second, we evaluated different architectures and post-processing \linebreak through user studies on two different datasets.

\edit{An earlier version of this work was presented in \citet{kucherenko2019analyzing}. This article extends the prior publication by considering an additional baseline model, by evaluating on a new dataset, by adding a more detailed numerical analysis, and by performing more in-depth user studies in two different languages.}


\section{Related work}
\label{sec:rel_work}

While most previous work on non-verbal behavior generation considers rule-based systems \citep{cassell2001beat,huang2012robot,salvi2009synface}, we here review only data-driven approaches and pay special attention to methods incorporating elements of representation learning, since that is an important direction of our research. For a review of rule-based systems we refer the reader to \citet{wagner2014gesture}.

\subsection{Data-driven head and face movements}
\label{ssec:dd_facial}

\upd{Facial gestures share many characteristics with hand gestures (for example high variability and weak correlation with the input) and similar approaches have been shown to perform well in generating both gesture types. For example, similar normalizing-flow techniques have been used for both hand-gesture generation \citep{alexanderson2020style} and facial gesture generation \citep{jonell2020letsfaceit}.}
Several recent works have applied neural networks in this domain, e.g., \citet{haag2016bidirectional,greenwood2017predicting,sadoughi2018novel,suwajanakorn2017synthesizing}. Among the cited works, \citet{haag2016bidirectional} used a bottleneck network to learn compact representations, although their bottleneck features subsequently were used to define prediction inputs rather than prediction outputs as in the work we present. Our proposed method works on a different aspect of non-verbal behavior that co-occurs with speech, namely generating body motion driven by speech.  
 
\subsection{Data-driven body-motion generation}
\label{ssec:dd_gestures}

Generating body motion 
is an active area of research with applications to animation, computer games, and other simulations. 
Current state-of-the-art approaches in such body-motion generation are generally data-driven and based on deep learning \citep{zhang2018mode,li2017auto,pavllo2018quaternet}. \citet{zhang2018mode} generated impressive-looking quadruped motion based on mode-adaptive neural networks, which are neural networks whose weights are linear functions of the output of a second ``gating'' neural network. 
\citet{li2017auto} proposed a modified training regime to make recurrent neural networks generate human motion with greater long-term stability, while \citet{pavllo2018quaternet} formulated separate short-term and long-term recurrent motion predictors, using quaternions to more adequately express body rotations. 

Some particularly relevant works for our purposes are \citet{liu2014feature,holden2015learning,holden2016deep,butepage2017deep}. All of these leverage representation learning (various forms of autoencoders) that predict human motion, yielding accurate yet parsimonious predictors. \citet{habibie2017recurrent} extended this general approach to include an external control signal in an application to human locomotion generation with body speed and direction as the control input. Our approach is broadly similar, but generates body motion from\linebreak speech rather than position information.

\subsection{Speech-driven gesture generation}
\label{ssec:sd_gestures}

Speech-driven gesture generation differs from other body-motion generation tasks in that the control signal input is computed from speech.
Like body motion in general, gesture generation has also begun to shift towards data-driven methods, for example \citet{levine2010gesture,chiu2015predicting,hasegawa2018evaluation}. Several researchers have studied hybrid approaches that combine data-driven approaches with rule-based systems. For example, \citet{bergmann2009GNetIc} learned a Bayesian decision network for generating iconic gestures.
\citet{sadoughi2017speech} used probabilistic graphical models with an additional hidden node to provide contextual information, such as a discourse function. The produced gestures were smoothed by interpolation between keypoints.
They experimented on only three hand gestures and two head motions. We believe that regression methods that learn and predict arbitrary movements, like the one we are proposing, represent a more flexible and scalable approach than the use of discrete and pre-defined gestures.

The work of \citet{chiu2011train} is of great relevance to our present work, in that they took a regression approach and also utilized representation learning. Specifically, they used wrist height in upper-body motion to identify gesticulation in motion-capture data of persons engaged in conversation. A network based on Restricted Boltzmann Machines (RBMs) was used to learn representations of arm gesture motion, and these representations were subsequently predicted based on prosodic speech-feature inputs using another network also based on RBMs. They also smoothed the produced gestures by finding accelerations above a threshold and reducing the speed in those time-frames.
\citet{levine2010gesture} also used an intermediate state between speech and gestures. The main differences are that they used hidden Markov models, whose discrete states have less expressive power than recurrent neural networks, and that they selected motions from a fixed library, while our model can generate unseen gestures. Later, \citet{chiu2015predicting} proposed a method to predict co-verbal gestures using a machine-learning setup with feedforward neural networks followed by Conditional Random Fields (CRFs) for temporal smoothing. They limited themselves to a set of 12 discrete, pre-defined gestures and used a classification-based approach.

\new{Recently, several models utilizing adversarial \edit{losses or loss terms in nonverbal behavior generation} have been proposed \upd{for beat gesture generation} \citep{ginosar2019learning,ferstl2019multi}. \citet{ginosar2019learning} developed a model for 2D gesture generation, which is a significantly simpler task where one can utilize vast amounts of data from online videos, which is not available in 3D. Similar to the work in this paper, \citet{ferstl2019multi} used recurrent neural networks, but without representation learning. They did not perform any subjective evaluations, so it is unclear how natural their generated motion was. We therefore did not take these models as our baseline.

In another recent work, \citet{ahuja2019react} proposed a deep-learning based model for dyadic conversation. This requires one to know the behavior of the conversation partner, in contrast to the setting of a single \edit{talker} considered in this paper.}

\citet{hasegawa2018evaluation} designed a speech-driven neural network capable of producing natural-looking 3D motion sequences. They had to apply a smoothing filter to the produced motion in order to remove jerkiness. We built our model on this work while extending it with motion-representation learning, since learned representations have improved motion prediction in other applications, as surveyed in Sec.~\ref{ssec:dd_gestures}.

%

\section{Models for speech-motion mapping}
\label{sec:method}

\subsection{Problem formulation}
We frame the problem of speech-driven gesture generation as follows: given a sequence of speech features $\boldsymbol{s} = [\boldsymbol{s}_t]_{t=1:T}$ extracted from segments (frames) of speech audio at regular intervals $t$, the task is to generate a corresponding gesture sequence $\hat{\boldsymbol{g}} = [\hat{\boldsymbol{g}}_t]_{t=1:T}$ that a human might perform while uttering this speech.

The speech at time $t$ would typically be represented by some features $\boldsymbol{s}_t$, such as Mel-Frequency Cepstral Coefficients 
(MFCCs), commonly used in speech recognition, or prosodic features including pitch (F0), energy, and their derivatives, commonly used in speech emotion analysis.
Similarly, the ground-truth gestures $\boldsymbol{g}_t$ and predicted gestures $\hat{\boldsymbol{g}}_t$ are typically represented as sequences of 3D keypoint coordinates:
$ \boldsymbol{g}_t = $\linebreak $[x_{i,t}, y_{i,t}, z_{i,t}] _{i=1:n}\text{,}$,
where $n$ is the number of keypoints on the human body (e.g., shoulder, elbow, etc.). 

Many recent gesture-generation systems perform the mapping from $\boldsymbol{s}$ to $\hat{\boldsymbol{g}}$ using a neural network (NN) learned from data; see Sec.~\ref{sec:rel_work}. The dataset typically contains recordings of human motion (for instance from motion capture) and the concurrent speech signals.

\begin{figure}
\centering
\vspace{-1mm}
\subfloat[Aud2Pose\label{sfig:Aud2pose}]{\includegraphics[width=0.43\linewidth]{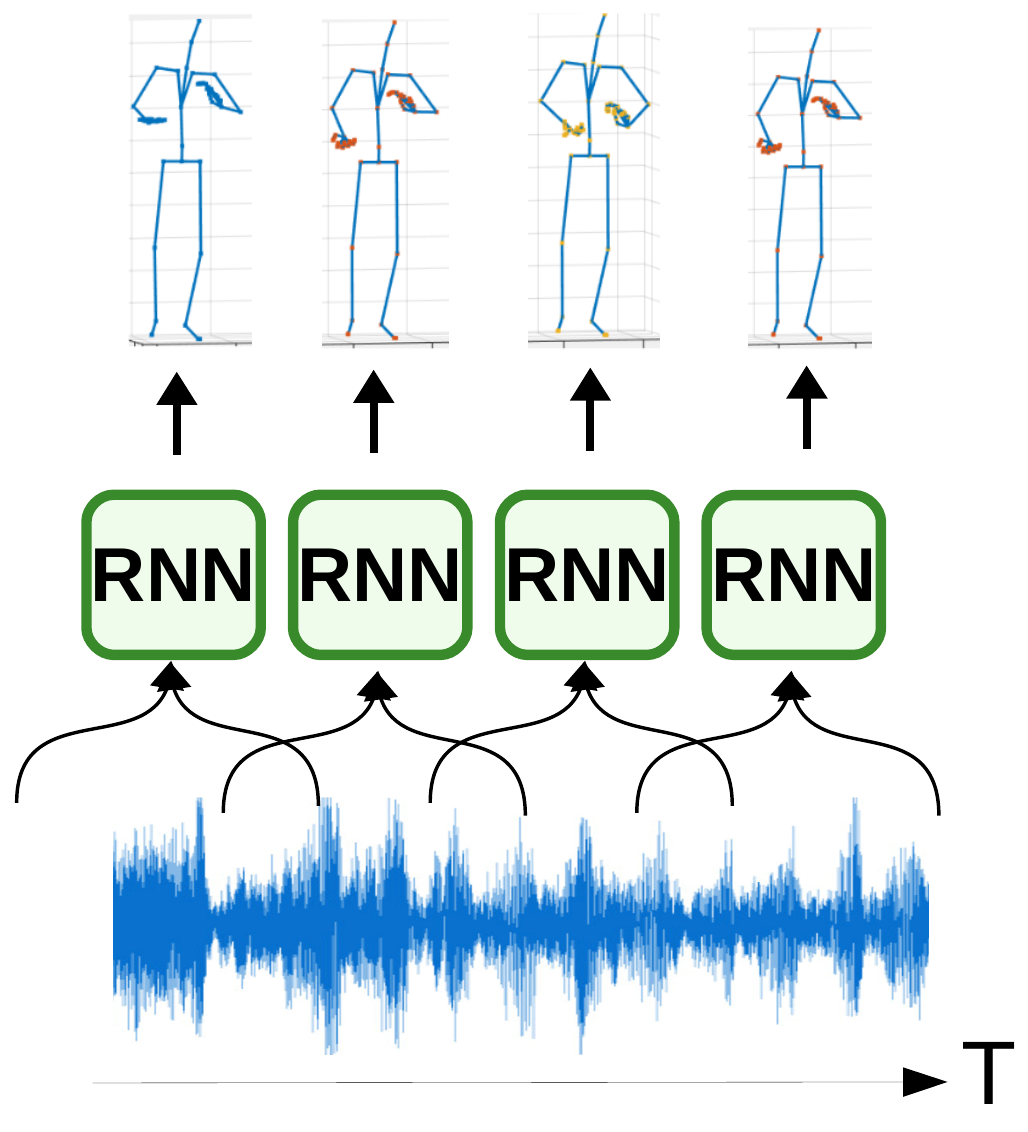} }
\subfloat[Aud2Motion\label{sfig:Aud2motion}]{\includegraphics[width=0.39\linewidth]{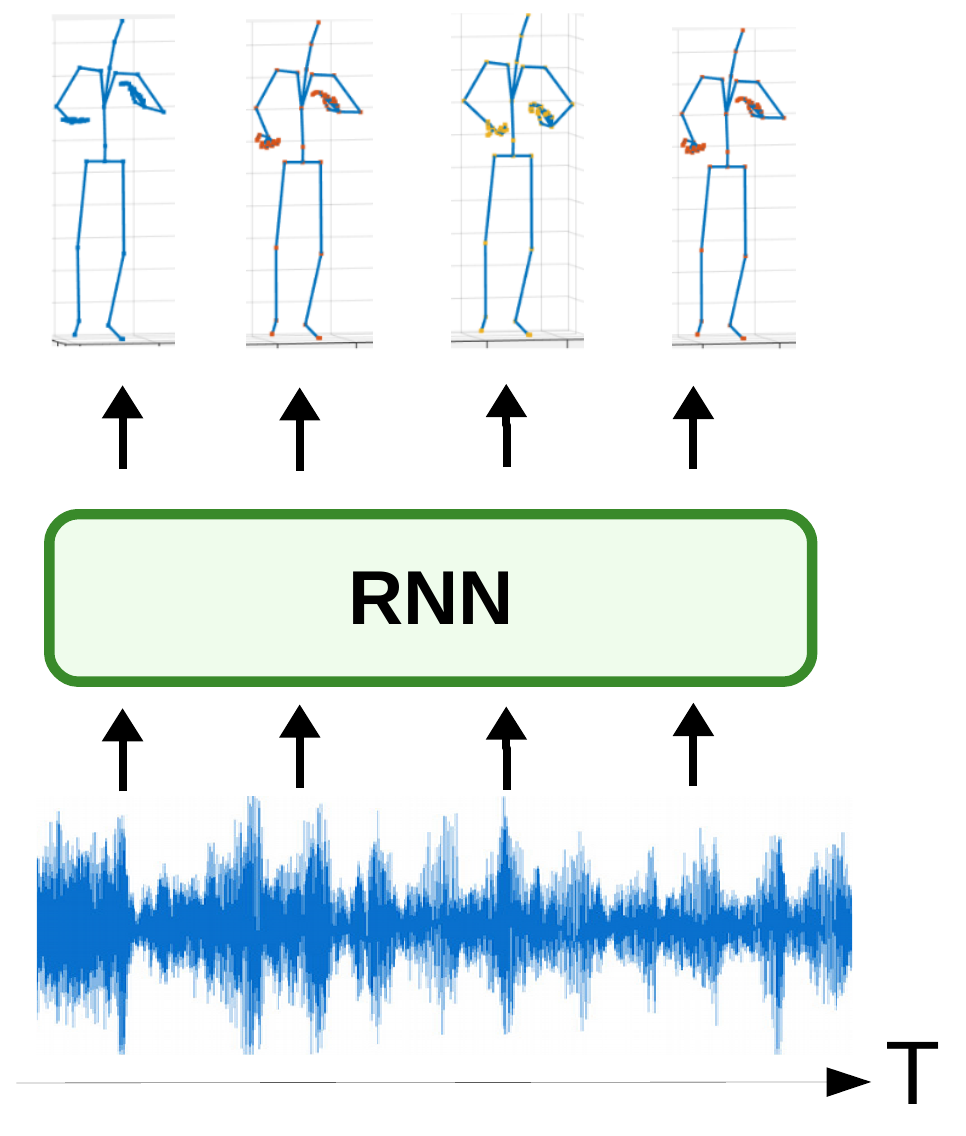}}
\vspace{-1mm}
\caption{Two different approaches to speech-motion mapping. Aud2Pose model maps from audio sequence to a motion frame, while Aud2Motion model maps audio sequence to a motion sequence instead. The figure exaggerates the offset between frames for demonstration purposes.}
\label{fig:framework_overview}
\vspace{-3mm}
\end{figure}

\subsection{Speech-to-motion models}
\label{ssec:baseline}

The models we consider build on the work of \citet{hasegawa2018evaluation}. First, we briefly describe their model, which we will denote \textit{Aud2Pose} (for \emph{audio to pose}) and use as the baseline system for our work.
The Aud2Pose neural network in \citet{hasegawa2018evaluation} takes a window of speech-feature frames as input in order to generate a single frame of motion. By sliding the window over the speech input sequence, a sequence of predicted poses is produced. As illustrated in Fig.~\ref{sfig:Aud2pose}, the speech windows contain chunks of $C=30$ frames before and after the current time $t$
. 
At each time step $t$ the entire window is fed as input to the network: $\mathrm{NN}_{\mathrm{input}} = [\textbf{s}_{t-C}, ... \textbf{s}_{t-1}, \textbf{s}_t, \textbf{s}_{t+1}, \ldots \textbf{s}_{t+C}]$.
The network is regularized by predicting not only the pose but also the velocity as output: $\mathrm{NN}_{\mathrm{output}} = [\textbf{g}_{t},\Delta \textbf{g}_{t}]$. While incorporating the velocity (finite difference) into predictions did not provide a significant improvement at test time, including velocity as a multitask objective helped the network learn motion dynamics during training \citep{takeuchi2017speech}.

As seen in \new{Fig.~\ref{sfig:Aud2pose}}, the baseline network (Aud2Pose) can be categorized as a many-to-one network, which takes audio features of multiple frames as input and outputs a single frame of gesture.
One may be curious about a many-to-many version of the network, which should  be able to take the temporal consistency of both inputs and outputs into account.
Thus, we also implement such a network, which we call \textit{Aud2Motion}, \rev{as an additional baseline.}
The network structure is almost the same as the baseline except for the recurrent layer, which we modified to output the full sequence instead of output just one frame.
Consequently, the network accepts an arbitrary number of input frames and returns one output frame for every input frame, without using the sliding window. \upd{This setup means that the output pose at any given time depends on all preceding speech, instead of being restricted to only consider the speech inside a finite-length window.} The difference between the models is shown in \new{Fig.~\ref{fig:framework_overview}}.

\begin{figure}
\centering
\includegraphics[width=0.98\linewidth]{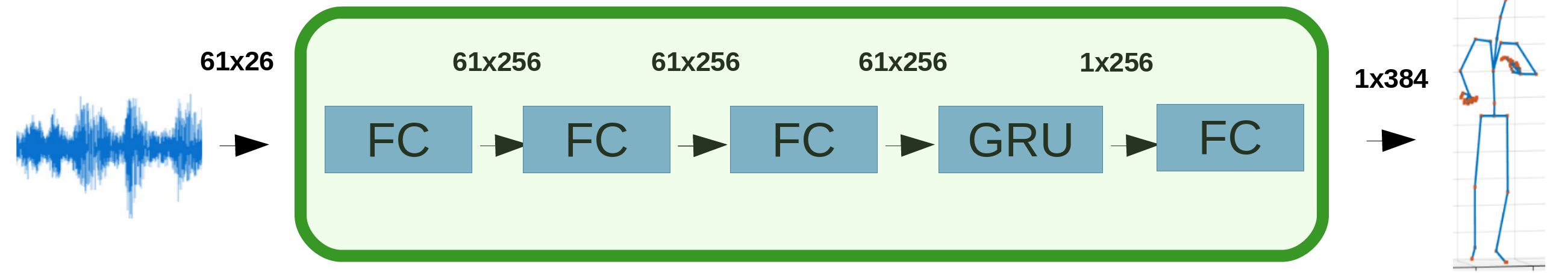}
\caption{Baseline RNN for speech-to-motion mapping. The green outline encloses the network labelled ``RNN'' in Fig.~\ref{fig:framework_overview}.}
\label{fig:baseline_model}
\vspace{-5mm}
\end{figure}

\new{Fig.~\ref{fig:baseline_model}} shows the neural network architecture used in this study. We use the same architecture for all systems in this paper to facilitate a fair comparison. The network input 
is passed through three Fully-Connected layers (FC). 
After that, a recurrent network layer with Gated Recurrent Units (GRUs) \citep{cho2014properties} is applied to the resulting sequence.
These GRUs were bidirectional in all systems except \textit{Aud2Motion}, where the GRU was unidirectional and causal.
Finally, an additional linear and fully-connected layer is used as the output layer.

We note that the baseline network we use is a minor modification of the network in \citet{hasegawa2018evaluation}. Specifically, we use a different type of recurrent network units, namely GRUs instead of B-LSTMs. Our experiments found that this cuts the training time in half while maintaining the same prediction performance.

\subsection{Proposed approach}
\label{ssec:proposed}

Our intent is to extend and improve the baseline model (\textit{Aud2Pose}) by leveraging the power of representation learning. \edit{We learn a more compact representation
by removing redundancy while preserving important information.
This simplifies learning, which is beneficial since \upd{synchronized data of speech audio together with 3D gestures is scarce}.}
\edit{We label our proposed approach \textit{Aud2Repr2Pose}. It} has three steps:
\begin{enumerate}
    \item We apply representation learning to learn a pose and velocity representation $\boldsymbol{z}$.
    \item We learn a mapping from the chosen speech features $\boldsymbol{s}$ to the learned representation $\boldsymbol{z}$ (using the same NN architecture as in the baseline model).
    \item The two learned mappings are chained together to turn speech input $\boldsymbol{s}$ into motion output $\hat{\boldsymbol{g}}$.
\end{enumerate}

\begin{figure}[t]
\centering
\subfloat[MotionED: Representation learning for the motion.\label{sfig:motioned}]{\hfill\includegraphics[width=0.73\linewidth]{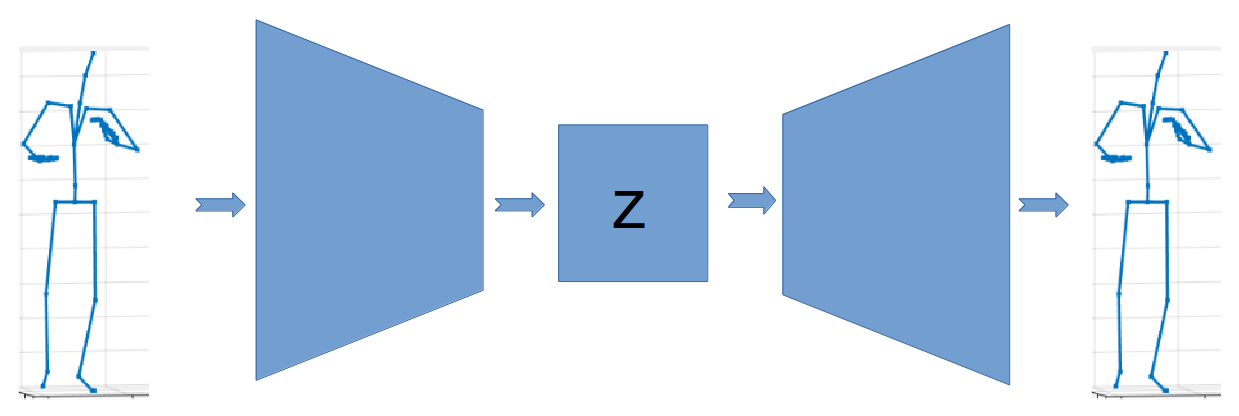}\hfill}\\
\vspace{-3mm}
\subfloat[SpeechE: Mapping speech to motion representations.\label{sfig:speeche}]{\hfill\includegraphics[width=0.73\linewidth]{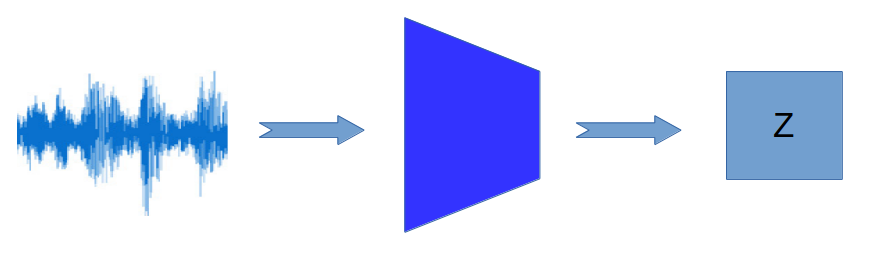}\hfill}\\
\vspace{-1mm}
\subfloat[Combining the learned components: SpeechE and MotionD.\label{sfig:combined}]{\hfill\includegraphics[width=0.73\linewidth]{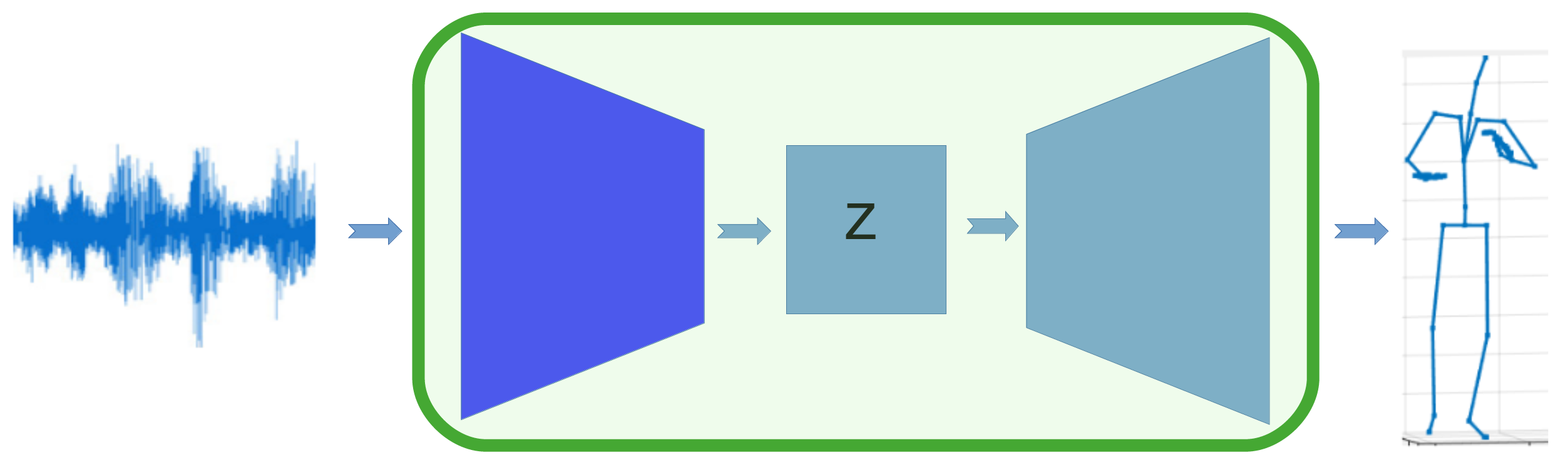}\hfill}
\vspace{-1mm}
\caption{Aud2Repr2Pose: Proposed encoder-decoder architecture for speech-to-motion mapping.}
\label{fig:our_model}
\end{figure}

\subsubsection{Motion-representation learning}

Fig.~\ref{sfig:motioned} illustrates representation learning for human motion sequences. The aim of this step is to reduce the motion dimensionality, which confers two benefits: 1) it simplifies the learning problem by reducing the output space dimensionality; and 2) it reduces redundancy in the training data by forcing the system to concentrate important information to fewer numbers.

To learn motion representations, we used a neural network structure called a Denoising Autoencoder (DAE) \citep{vincent2010stacked} with one hidden layer ($\boldsymbol{z}$).
This network learns to reconstruct the original data from input examples with additive noise while having a bottleneck layer in the middle. This bottleneck forces the network to compute a lower-dimensional representation.
The network can be seen as a combination of two sub-networks: \textit{MotionE}, which encodes the motion $\boldsymbol{m}$ to the representation $\boldsymbol{z}$ and \textit{MotionD}, which decodes the representation $\boldsymbol{z}$ back to the motion space $\hat{\boldsymbol{m}}$, i.e., 
$\boldsymbol{z} = MotionE(\boldsymbol{m})$, 
$\hat{\boldsymbol{m}} = MotionD(\boldsymbol{z})$.
The neural network learns to reconstruct the original motion input as closely as possible by minimizing the mean squared error (MSE) reconstruction loss: $\mathrm{MSE}(\boldsymbol{m},\hat{\boldsymbol{m}}) = \frac{1}{T} \lVert \hat{\boldsymbol{m}} - \boldsymbol{m}\rVert_2^2$.

\subsubsection{Encoding speech to the motion representation}

Fig.~\ref{sfig:speeche} illustrates the principle of how we map speech to motion representations. Conceptually, the network performing this task fulfills the same role as the baseline network in Sec.~\ref{ssec:baseline}. 
The main difference versus the baseline is that the output of the network is not raw motion values, but a compact learned representation of motion.
To be as comparable as possible to the baseline, we use the same network architecture to map speech to motion representations in the proposed system as the baseline used for mapping speech to motion. We call this network \textit{SpeechE}.

\subsubsection{Connecting the speech encoder and the motion decoder}

Fig.~\ref{sfig:combined} illustrates how the system is used at test time by chaining together the two previously learned mappings. First, speech input is fed to the \textit{SpeechE} transcoder, which produces a sequence of motion representations. Those motion representations are then decoded into joint coordinates using the \textit{MotionD} decoder.

\section{Experimental setup}
\label{sec:exp_setup}

This section describes the data and gives technical detail regarding the experiments we conducted to evaluate the importance of input and output representations in speech-driven gesture generation as well as the effect of motion post-processing, especially smoothing.

\subsection{Gesture-speech datasets}
\label{ssec:dataset}

For most of our experiments, we used a gesture-speech dataset collected by \citet{takeuchi2017creating}\footnote{\new{The dataset is available at \href{https://bit.ly/2Q4vSnT}{https://bit.ly/2Q4vSnT}}.}, from two Japanese individuals recorded separately in a motion-capture studio. The recordings had the form of an interview, where an experimenter asked questions prepared beforehand and a performer (the person being recorded) answered them. The dataset contains MP3-encoded speech audio captured using headset microphones on each performer, coupled with motion-capture motion data stored in the BioVision Hierarchy format (BVH). The BVH data describes motion as a time sequence of Euler rotations for each joint in the defined skeleton hierarchy. 
These Euler angles were converted to a total of 64 global joint positions in 3D\footnote{\edit{\citet{hasegawa2018evaluation} describe that they hip-centered the data as a pre-processing step, but we confirmed with the first author that the data was not actually hip-centered for system training.}}. As some recordings had a different framerate than others, we downsampled all recordings to a common framerate of 20 frames per second (fps).



This dataset contains 1,047 utterances\footnote{The original paper reports 1,049 utterances, which is a typo.}, of which our experiments used 957 for training, 45 for validation, and 45 for testing.  
The relationship between various speech-audio features and the 64 joint positions was thus learned from 171 minutes of training data at 20 fps, resulting in \edit{206k} training frames.

For our additional evaluation, we used a gesture-speech dataset collected by \citet{ferstl2018investigating}. Motion data were recorded in a motion-capture studio from an Irish actor speaking impromptu with no fixed topic. This dataset also contains audio in MP3 format coupled with motion-capture data in FBX format. To be maximally comparable with the Japanese dataset, we downsampled all these recordings to 20 fps.
As before, the Euler angles were converted to global joint positions in 3D.
We consider only upper body motion, so this data has fewer joints: 46. 
This dataset contains 23 recordings, each on average ten minutes long. For our experiments we used 19 of them for training, 2
for validation, and 2 for testing.  The training dataset comprised 202 minutes at 20 fps, resulting in \edit{243k} frames.

For the representation learning, each dimension of the motion was standardized to mean zero and maximum (absolute) value one.

\subsection{Feature extraction}
\label{sec:technical_details}

The ease of learning and the limits of expressiveness for a speech-to-gesture system depend greatly on the input features used. Simple features that encapsulate the most important information are likely to work well for learning from small datasets, whereas rich and complex features might allow learning additional aspects of speech-driven gesture behavior, but may require more data to achieve good accuracy. We experimented with three different, well-established audio-feature \edit{representations} as inputs to the neural network, namely i) MFCCs, ii) spectrograms, and iii) prosodic features.


In terms of implementation, 26 MFCCs were extracted with a window length of 0.02 s and a hop length of 0.01 s, which amounts to 100 analysis frames per second. \new{Note that we used a shorter window length than the baseline paper \citet{hasegawa2018evaluation} (i.e., 0.02 s instead of 0.125 s), as MFCCs were developed to be informative about speech for these window lengths.} Our spectrogram features, meanwhile, were 64-dimensional and extracted using Librosa \citep{mcfee2015librosa} with hop length 5 ms and default window length (here 46 ms), yielding a rate of 200 fps. Frequencies that carry little speech information (below the hearing threshold of 20 Hz, or above 8000 Hz) were removed. Both the MFCC and the spectrogram sequences were downsampled to match the motion frequency of 20 fps by replacing every 5 (MFCCs) or 10 (spectrogram) frames by their average. 

As an alternative to MFCCs and spectrum-based features, we also considered prosodic features. These differ in that prosody encompasses intonation, rhythm, and anything else about the speech outside of the specific words spoken (e.g., semantics and syntax). Prosodic features were previously used for gesture prediction in early data-driven work by \citet{chiu2011train}. For this study, we considered pitch and energy (intensity) information. The information in these features has a lower bitrate and is not sufficient for discriminating between and responding differently to arbitrary words, but may still be informative for predicting non-verbal emphases like beat gestures and their timings.

We considered four specific prosodic features, extracted from the speech audio \new{with a hop length of 5 ms}, resulting in 200 fps. Our two first prosodic features were the energy of the speech signal and the time derivative (finite difference) of the energy series. The third and fourth features were the logarithmically transformed F0 (pitch) contour, which contains information about the speech intonation, and its time derivative. The pitch in unvoiced frames was set to 0 Hz. We extracted pitch and intensity values from audio \new{using Praat \citep{boersma2002praat} with default settings} and normalized pitch and intensity as in \citet{chiu2011train}: pitch values were adjusted by taking $\log(x+1)-4$ and setting negative values to zero, and intensity values were adjusted by taking $\log(x)-3$. All these features were again downsampled to the motion frequency of 20 fps using averaging. 

 

\subsection{Implementation details}

To facilitate reproducing our results, our code is publicly available.%
\footnote{See \href{https://github.com/GestureGeneration/Speech_driven_gesture_generation_with_autoencoder}{https://github.com/GestureGeneration/Speech\_driven\_gesture\_generation\_with\_autoencoder}}



\subsubsection{The Aud2Pose neural network}

The fully-connected layers and the GRU layers both had a width of 256 nodes and used the ReLU activation function. Batch normalization and dropout with probability 0.1 of dropping activations were applied between every layer. Training minimized the mean squared error (MSE) between predicted and ground-truth gesture sequences using the Adam optimizer \citep{kingma2014adam} with learning rate 0.001 and batch size 2056. Training was run for 120 epochs, after which no further improvement in validation loss was observed. \new{All the hyperparameters except batch size and number of epochs (which were adjusted based on the validation accuracy) were taken from the baseline paper} \citet{hasegawa2018evaluation}.

\subsubsection{The denoising autoencoder (MotionED) neural network}


We trained a DAE with input size 384 (64 joints: 192 3D coordinates and their first derivatives) and one hidden, feedforward layer in the encoder and decoder. 
Hyperparameters were optimized on our validation dataset, described in Sec.~\ref{ssec:dataset}.
Different widths were investigated for the bottleneck layer (see Sec.~\ref{ssec:dim_analysis}), with 325 units giving the best validation-data performance. During training, Gaussian noise was added to each input with a standard deviation equal to 0.05 times the standard deviation of that feature dimension. Training then minimized the MSE reconstruction loss using Adam with a learning rate of 0.001 and batch size 128. Training was run for 20 epochs.

\subsubsection{The Aud2Motion neural network}

While the Aud2Motion approach can accept an arbitrary number of frames as inputs, we partitioned training sequences into fixed-length chunks to make training time reasonable and cope with the gradient vanishing problem.
We empirically set the chunk length to 100 thus the network inputs had 100 $\times$ 26 elements (100 time-steps of MFCC audio features).
The outputs of the network also contained 100 frames.
Note that adding velocity to the expected network output as regularization (described in Sec.~\ref{ssec:baseline}) was omitted for the Aud2Motion since the multi-frame outputs already had information about the velocity. (In fact, we found no performance improvements with the regularization in the Aud2Motion.)
All hyperparameters but the number of epochs (set to 500) were the same as the baseline.
At testing time, we fed variable-length inputs to the network instead of the fixed-length chunks to produce continuous predictions over full input sequences.


\subsubsection{Adjustments for the second dataset}

We have tuned network parameters based on validation accuracy using random search, which found that different models had different optimal hyperparameters on different datasets. The following aspects were changed for the experiments on the second dataset:
1) On this data, we did not predict velocities, but only coordinates.  
2) for the Aud2Repr2Pose model the representation size was reduced to 118; 
3) for the Aud2Motion model the batch size was 256.



\subsection{Numerical evaluation measures}
\label{ssec:evaluation_measures}

We used both objective and subjective measures to evaluate the different approaches under investigation.
Among the former, two kinds of error measures were considered:
\begin{description}
\item \textit{Average Position Error (APE).}
The APE is the average Euclidean distance between the predicted coordinates $\hat{g}$ and the original coordinates $g$. \new{For each sequence $n$:
\begin{equation}
\label{eq:task}
\mathrm{APE}(g^n,\hat{g}^n) = \frac{1}{KT} \sum_{t=1}^T \sum_{k=1}^K \lVert g_{kt}^n - \hat{g}_{kt}^n \rVert_2
\end{equation}
where $T$ is the duration of the sequence, $K$ is the number of joints and $g_{kt}$ is the Cartesian 3D coordinates of joint $k$ in frame $t$.}
\item \textit{Motion Statistics.}
We considered the average values and distributions of speed and jerk for the generated motion.
\new{The speed between time $t$ and $t-1$ is computed by taking a finite difference between joint positions at time $t$ and $t-1$.
\emph{Jerk} is a temporal differentiation of acceleration (i.e., it is the third derivative of joint position) and quantifies the smoothness of the motion.
Smoother gesture motion has lower jerk.}
\end{description}

We believe the motion statistics to be the most informative objective measures for our task: in contrast to tracking (3D pose estimation), the purpose of gesture generation is not to reproduce one specific \textit{true} position, but rather to produce a plausible candidate for natural motion. Plausible motions do not require measures like speed or jerk to closely follow the original motion, but they should produce a similar distribution. That is why we study distribution statistics, namely \new{velocity, acceleration}, and jerk.

Since there is some randomness in system training, e.g., due to random initial network weights, we trained systems for every condition five times and report the mean and standard deviation of those results.
\section{Numerical Evaluation with Discussion}
\label{sec:exper_res}
This section presents a numerical analysis of the performance of the gesture-prediction systems.
We investigated different aspects of system design that relate to the importance of representations of speech and motion and the post-processing of the gestures.

\begin{figure}
\centering
\subfloat[Average position error (APE). The baseline APE (blue line) is 8.3$\mathbf{\pm}$0.4 \edit{cm}.\label{sfig:ape}]{\includegraphics[width=0.85\columnwidth]{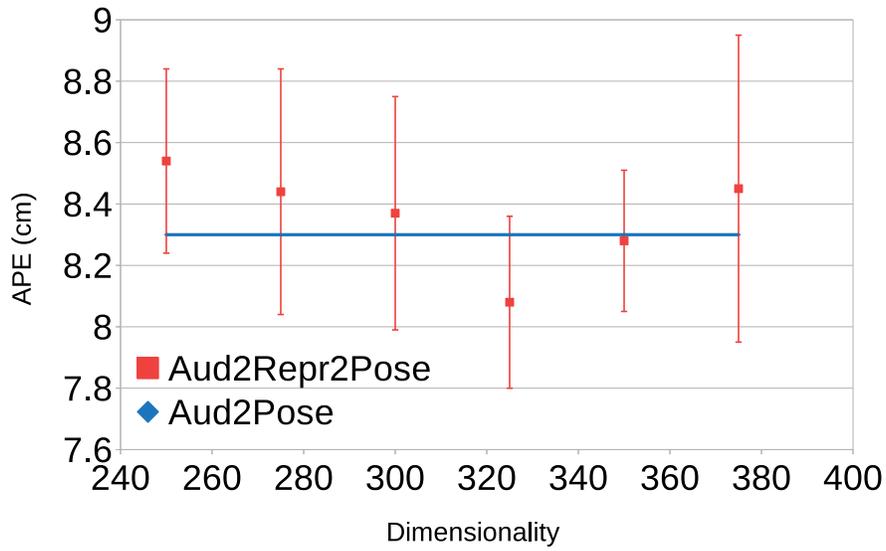}}
\hfill
\subfloat[Average jerk. Baseline jerk is \edit{ 56$\mathbf{\pm}$6 cm/s$^3$ while ground-truth jerk is 10.8 cm/s$^3$}.\label{sfig:avg_jerk}]{\includegraphics[width=0.85\columnwidth]{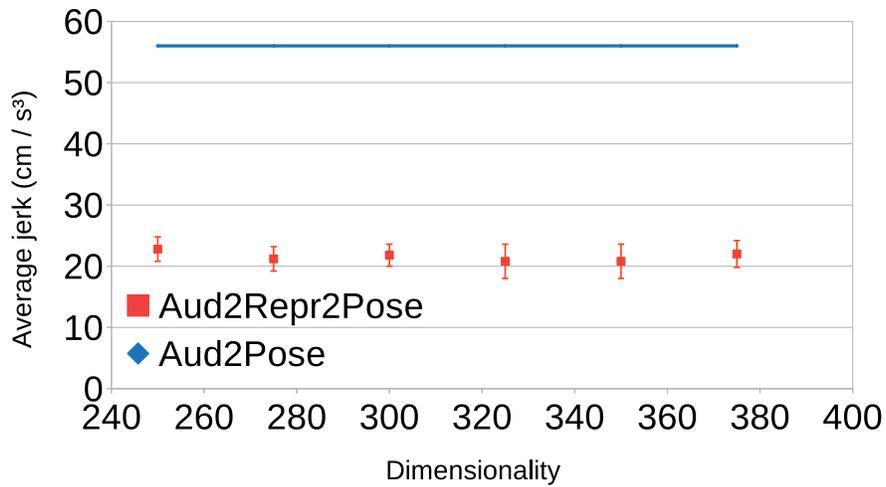}}\\
\caption{Effect of learned-representation dimensionality in the proposed model.}
\label{fig:dim_analysis}
\end{figure}

\subsection{Importance of motion encoding}
\label{ssec:dim_analysis}

We first evaluated how different dimensionalities for the learned motion representation affected the prediction accuracy of the full system.
Fig.~\ref{fig:dim_analysis} graphs the results of this evaluation. In terms of the average position error (APE) in Fig.~\ref{sfig:ape}, the optimal embedding-space dimensionality is 325, which is smaller than the original data dimensionality (384). Motion jerkiness (see Fig.~\ref{sfig:avg_jerk}) is also lowest for dimensionality 325, but only by a slight margin compared to the uncertainty in the estimates. Importantly, the proposed system performs better than the baseline \citet{hasegawa2018evaluation} on both evaluation measures. The difference in average jerk, in particular, is highly significant. This validates our decision to use representation learning to improve gesture generation models.

The numerical results are seen to vary noticeably between different runs, suggesting that training might converge to different local optima depending on the random initial weights.

\subsection{Input-speech representation}

Having established the benefits of representation learning for the output motion, we next analyzed which input features performed the best for our speech-driven gesture generation system. In particular, we compared three different features \edit{and their combinations} -- MFCCs, raw power-spectrogram values, and prosodic features (log F0 contour, energy, and their derivatives) -- as described in Sec.~\ref{sec:technical_details}.

\edit{Table \ref{tab:speech_features} displays average values and standard deviations of the mean absolute speed and other motion statistics, calculated over several training runs of each model.
From the table}, we observe that MFCCs achieved the lowest APE, but produced motion with higher 
jerkiness than the spectrogram features did. Spectrogram features gave suboptimal APE, but match ground-truth 
jerk better than the other features.



\begin{table}
  \caption{Objective evaluation of different systems and speech features, averaged over five re-trainings of the systems. \edit{Each cell reports the mean values and standard deviations over five different training runs. The best values in each column are indicated in bold. APE stands for Average Position Error, Spd for Speed, Acc for acceleration, Spec for Spectrogram, and Pros for Prosodic features.}}
  \label{tab:speech_features}
  \begin{tabular}{@{}llcccc@{}}
    \toprule
    Model & Features & APE (\edit{cm}) & \edit{Spd (cm/s)} & \new{Acc} \edit{(cm/s$^2$)} & Jerk \edit{(cm/s$^3$)} \\
    \toprule
    {\small Aud2Repr2Pose} & {\small Prosodic} & 8.56 $\pm$ {0.2}  & \edit{4.8 $\pm$ {0.1}}  & \edit{18.0 $\pm$ {0.6}} & 30.4 $\pm$ {1.4}\\
    {\small Aud2Repr2Pose}  & {\small Spectrogram} &  8.27 $\pm$ {0.4}   & \edit{4.8 $\pm$ {0.1}} & \textbf{\edit{10.2}} $\pm$ {1.4}  & \textbf{17.0} $\pm$ {2.4}\\
    {\small Aud2Repr2Pose} & {\small Spec+Pros} & 8.11 $\pm$ {0.3}  &  \edit{4.4 $\pm$ {0.3}} & \edit{11.4 $\pm$ {1.6}}   & 19.0 $\pm$ {2.4}\\
    {\small Aud2Repr2Pose}& {\small MFCC} & \textbf{7.66} $\pm$ {0.2} & \edit{4.5 $\pm$ {0.2}} & \textbf{\edit{10.6}} $\pm$ {0.6}  & 18.2 $\pm$ {1.0} \\
    {\small Aud2Repr2Pose} & {\small MFCC+Pros} & \textbf{7.65} $\pm$ {0.2}  &  \edit{4.5 $\pm$ {0.1}}  & \edit{11.6} $\pm$ {1.2} & 19.4 $\pm$ {2.2}\\
    {\small Aud2Repr2Pose} & {\small \new{MFCC+Spec}} & \new{7.75} $\pm$ {0.1} & \edit{4.5 $\pm$ {0.2}}   &  \new{13.6} $\pm$ {0.6}  & \new{22.0} $\pm$ {1.0}\\
    {\small Aud2Repr2Pose} & {\small \new{All three}} & \textbf{\new{7.65}} $\pm$ {0.1}  & \edit{4.5 $\pm$ {0.1}}  & \new{13.2} $\pm$ {0.4}  & \new{21.2} $\pm$ {0.6}\\
    \midrule
    {\small Aud2Pose} & {\small MFCC} & 8.07 $\pm$ {0.1}  & \edit{8.3 $\pm$ {0.2}} & 30.0 $\pm$ {0.6}    & 52.4 $\pm$ {1.0}\\
    \midrule
    {\small Aud2Motion} & {\small MFCC} & 8.01 $\pm$ {0.16} & \edit{8.5 $\pm$ {0.1}}   & \new{27.8 $\pm$ {1.2}} & 48.6 $\pm$ {2.2} \\
    \midrule
    {\small Static mean pose} & & 8.95  & \new{0} & 0 & 0\\
    \midrule
    {\small Ground truth} & & 0 & \edit{3.9} & \edit{7.2} & 10.8
\end{tabular}
\end{table}
Among other results we note that the \textit{Aud2Motion} model did not improve on \rev{the \textit{Aud2Pose} model.} It has similar jerkiness to the \textit{Aud2Pose} model, which is much higher than that of the ground truth, indicating that the motion is less smooth than natural trajectories are. There can be several reasons why the sequence-level \textit{Aud2Motion} model did not learn to produce a smooth motion. \upd{An important difference between And2Pose and Aud2Motion is that Aud2Pose uses future speech, while Aud2Pose generates motion frame-by-frame and only has access to the past speech.
We believe this explains the large difference between the two.
The fact that Aud2Motion model failed to perform well indicates that future speech is indeed important for the gesture generation task},
as gestures often precede speech \citep{pouwQuantifying}. 

\edit{Apart from looking at the mean of the motion statistics across several runs (as tabulated in Table \ref{tab:speech_features}), we also studied the distributions of motion statistics across different joints and different time frames within a single run. We observe that all motion statistics, including those of the ground truth, exhibit high standard deviation across the board. The ground truth distribution of absolute speed (for all joints and all frames) has a standard deviation of 3.79, which is very similar to the mean absolute speed (3.9). Evidently, natural motion can be both fast and slow. The wide range of variation makes it difficult to draw any deep conclusions based on statistical averages like those in Table \ref{tab:speech_features}
and motivates us to perform a more detailed analysis of the motion statistics.}

\subsection{Detailed performance analysis}

The objective measures in Table \ref{tab:speech_features} do not unambiguously establish which input features would be the best choice for our model. 
We therefore further analyzed the statistics of the generated motion, particularly the speed. Producing the right motion with the right speed distribution is crucial for generating convincing motion, as too fast or too slow motion does not look natural.

\begin{figure}
\centering
\subfloat[Average speed histogram.\label{sfig:feat_avg_acc}]{\hfill\includegraphics[width=0.7\linewidth]{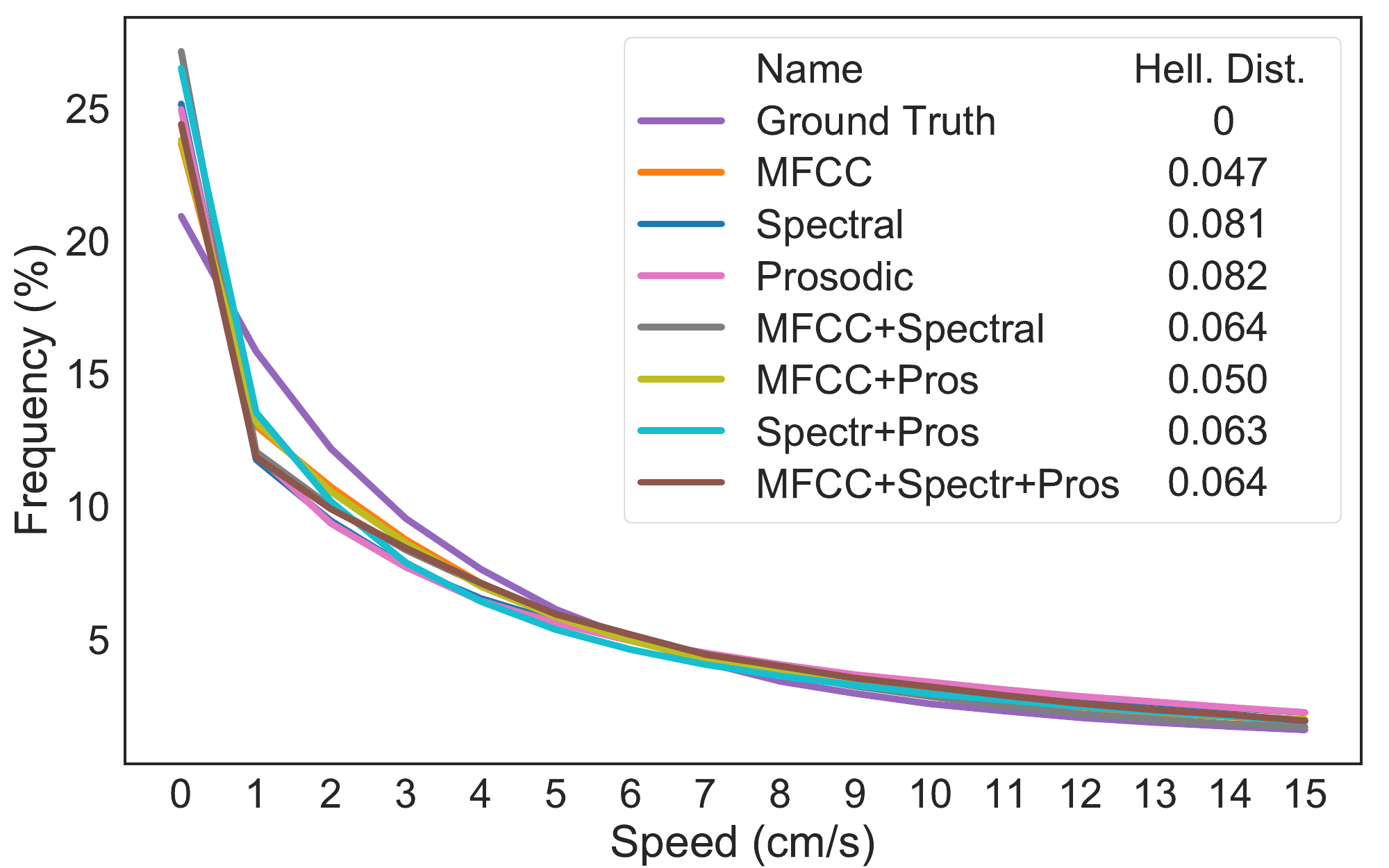}\hfill}\\
\subfloat[Speed histogram for forearm.\label{sfig:feat_elbows_acc}]{\hfill\includegraphics[width=0.7\linewidth]{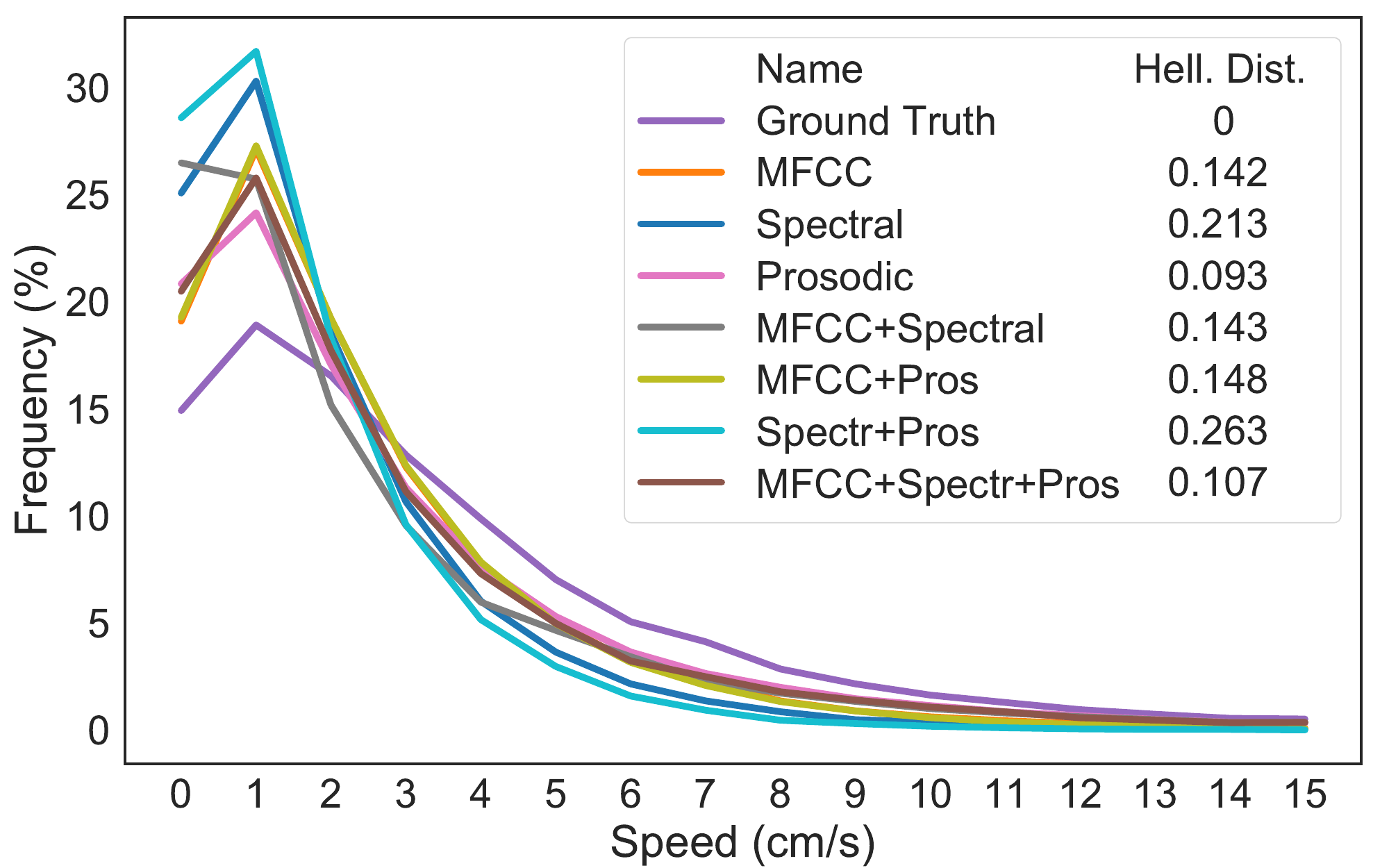}\hfill}\\
\subfloat[Speed histogram for wrists.\label{sfig:feat_hands_acc}]{\hfill\includegraphics[width=0.7\linewidth]{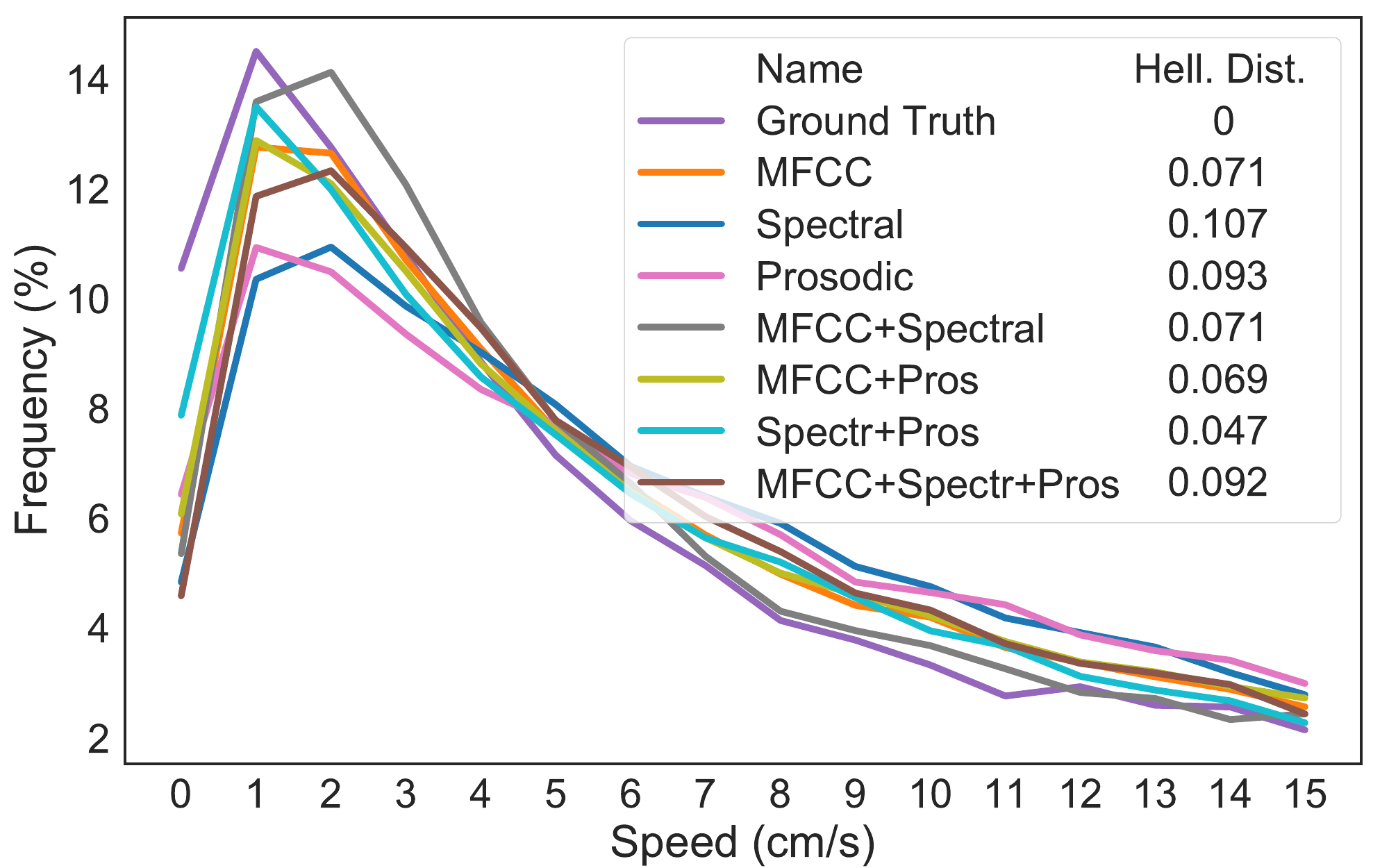}\hfill}\\
\caption{Speed distributions given different speech features. \new{No single feature choice is the best in all figures.}%
}
\label{fig:speed_hist_features}
\end{figure}

To investigate the motion statistics associated with the different input features, we computed speed histograms of the generated motions and compared those against histograms derived from the ground truth. We calculated the relative frequency of different speed values over frames in all 45 test sequences, \edit{and split the speed values} into bins of equal width. 
For easy comparison, our histograms are visualized as line plots rather than bar plots.  

\begin{figure}
\centering
\subfloat[Average speed histogram.\label{sfig:avg_acc}]{\hfill\includegraphics[width=0.65\linewidth]{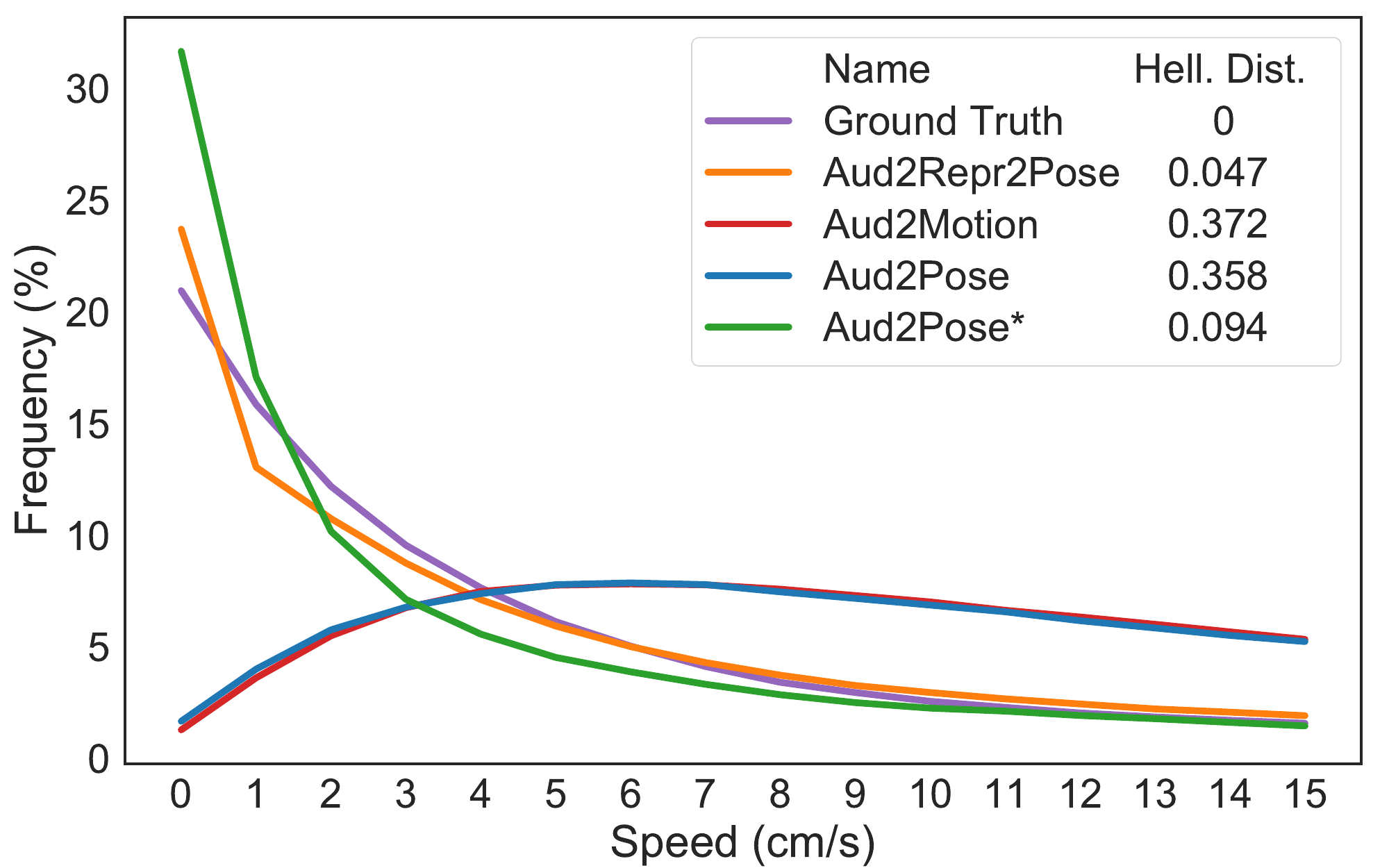}\hfill}\\
\subfloat[Speed histogram for elbows.\label{sfig:hands_acc}]{\hfill\includegraphics[width=0.65\linewidth]{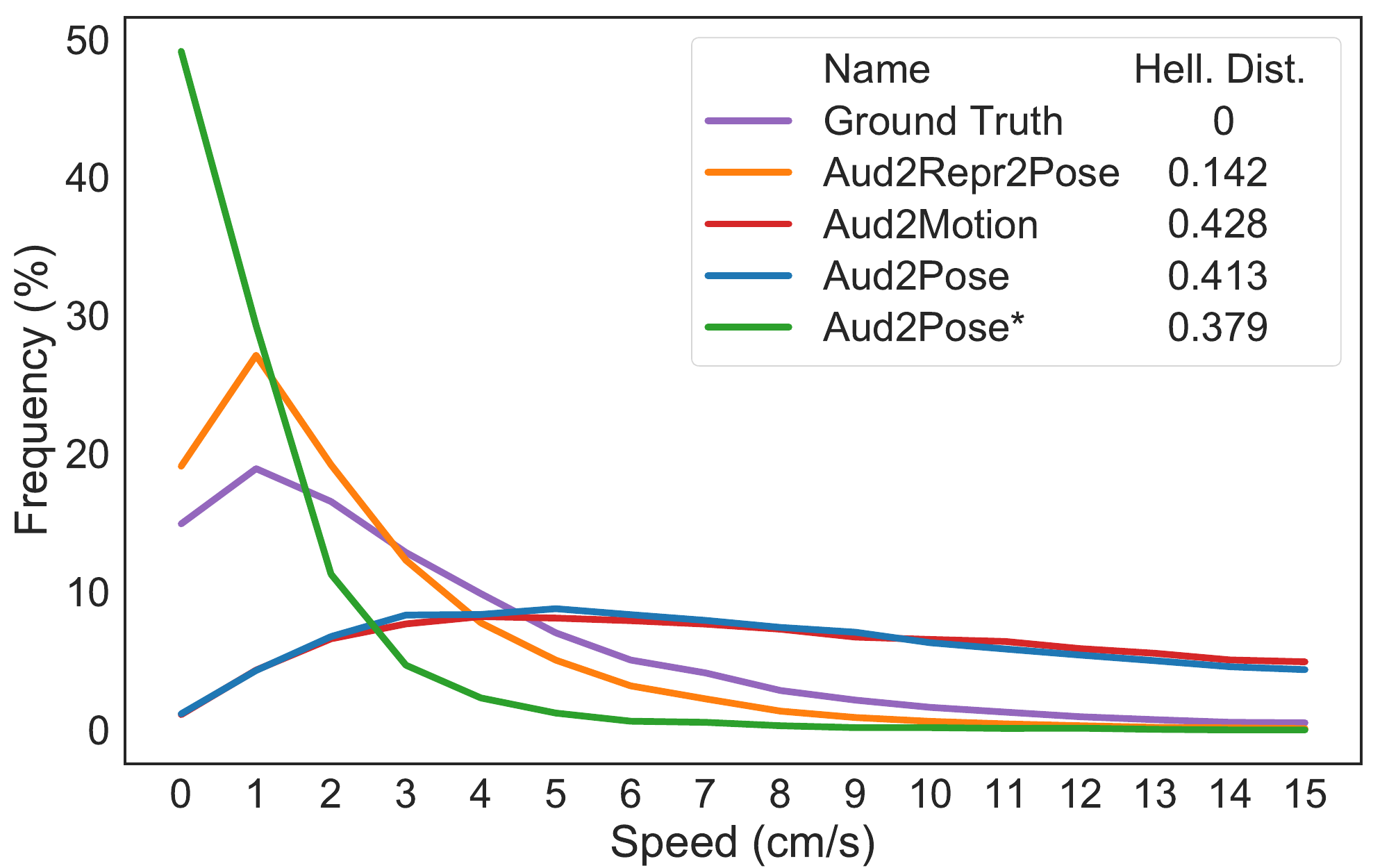}\hfill}\\
\subfloat[Speed histogram for wrists.\label{sfig:wrist_acc}]{\hfill\includegraphics[width=0.65\linewidth]{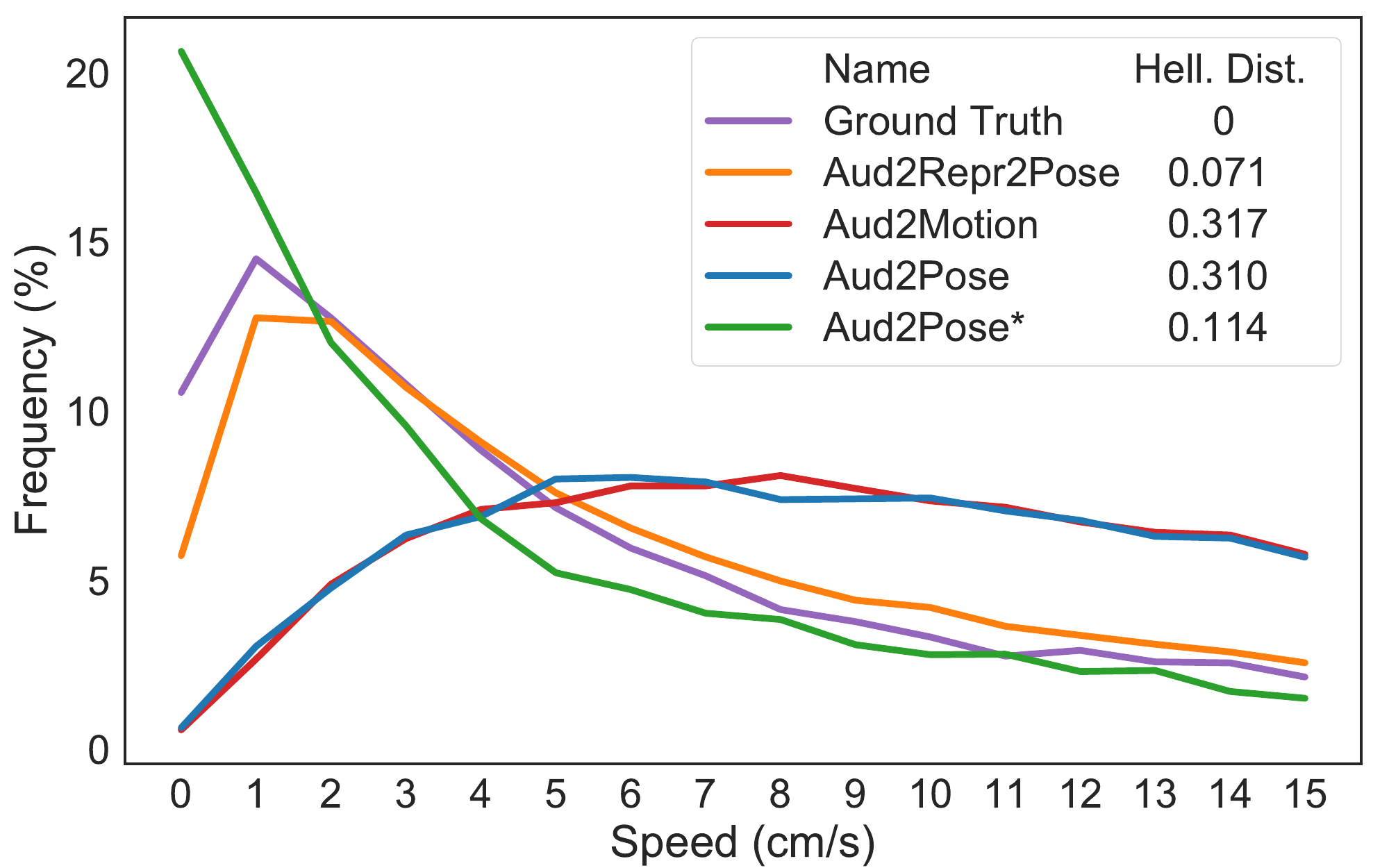}\hfill}\\
\caption{\new{Speed distributions for different models.
The motion produced from our model is more similar to the speed distribution of the ground-truth motion, compared to motion from the baseline model (Audio2Pose from \citet{hasegawa2018evaluation}). We can see that hip-centering and smoothing, denoted by a ``*'', together make a very substantial difference for the baseline model.}  }
\label{fig:speed_hist_models}
\end{figure}

Fig.~\ref{sfig:feat_avg_acc} presents speed histograms across all joints. 
\new{We observe no difference between using different features on this scale.} 
Since the results in Fig.~\ref{sfig:feat_avg_acc} are averaged over all joints, they do not indicate whether 
all the joints move naturally. To address this we also analyze the speed distribution for
the elbows (Fig.~\ref{sfig:feat_elbows_acc}) and wrists (Fig.~\ref{sfig:feat_hands_acc}). 
Hands convey the most important gesture information, suggesting that these plots are more informative. \new{To aid comparison we complement the visuals by computing a numerical distance measure between the different histograms and the ground truth.
We use the Hellinger distance $H$, which is a distance metric between two probability distributions.
For two normalized histograms $h^1$ and $h^2$, it is defined by:
\begin{equation}
  H(h^1, h^2) = \sqrt{1 - \sum_{i}{\sqrt{h^1_i \cdot h^2_i}}}.
\end{equation}
The calculated distances are provided in the legend for the corresponding figures.}

\new{Fig.~\ref{fig:speed_hist_models} presents speed histograms for different models using the same features (namely MFCCs).} 
The speed distributions of the \textit{Aud2Pose} and
\textit{Aud2Motion} models 
deviate much more from the ground truth than models with representation learning do. \new{A hypothetical explanation could be that the former models overfitted. However, their training curves show that the validation loss remains close to the training loss and does not increase, which contradicts the overfitting hypothesis.} Translating the motion to a hip-centered coordinate system and then smoothing it removed this deviation, a finding that we will discuss in greater depth in Sec.~\ref{sec:subj}.
 
 \subsection{Sequence-level comparisons}
 
 \upd{In the previous experiments we compared motion statistics aggregated over all output motion. However, different motions have different speeds, so these grand averages do not tell us whether synthetic motion is appropriately fast when natural motion is fast, and slow when natural motion is slow.
 For this reason we also analyze motion statistics in further detail by comparing speed histograms for each test sequence separately, and report on these statistics across the different sequences.}

\begin{table}
  \caption{\upd{Hellinger distances of speed distributions for different speech features, computed on a per-sequence basis. The table shows average values and standard deviations aggregated across all test sequences. In each column the values closest to the ground truth are highlighted in bold.}}
  \label{tab:speech_sequence_features}
  \centering
  \begin{tabular}{@{}lccc@{}}
    \toprule
    \multirow{2}{*}{Features} & \multicolumn{3}{c}{Hellinger distance w.r.t.\ ground truth} \\
     & All joints & Elbows & Wrists \\
    \toprule
    {\small Prosodic} & 0.133 $\pm$ 0.045 & \textbf{0.220} $\pm$ 0.113 & 0.214 $\pm$ 0.104 \\
    {\small Spectrogram} & 0.147 $\pm$ 0.052 & 0.287 $\pm$ 0.115 & 0.213 $\pm$ 0.115 \\
    {\small Spec+Pros} & 0.152 $\pm$ 0.062 & 0.274 $\pm$ 0.105 & 0.238 $\pm$ 0.134 \\
    {\small MFCC} & 0.133 $\pm$ 0.062 & 0.271 $\pm$ 0.147 & 0.214 $\pm$ 0.106 \\
    {\small MFCC+Pros} & 0.131 $\pm$ 0.057 & 0.286 $\pm$ 0.126 & 0.215 $\pm$ 0.102 \\
    {\small MFCC+Spec} & 0.132 $\pm$ 0.051 & 0.259 $\pm$ 0.104 & \textbf{0.206} $\pm$ 0.102 \\
    {\small All three} & \textbf{0.124} $\pm$ 0.053 & 0.246 $\pm$ 0.111 & 0.209 $\pm$ 0.100 \\
    \midrule
    {\small Ground truth} & 0 & 0 & 0 \\
    \bottomrule
\end{tabular}
\end{table}

\begin{figure}
\centering
\subfloat[Seq.~1152: Average speed histogram.\label{sfig:avg_acc}]{\hfill\includegraphics[width=0.45\linewidth]{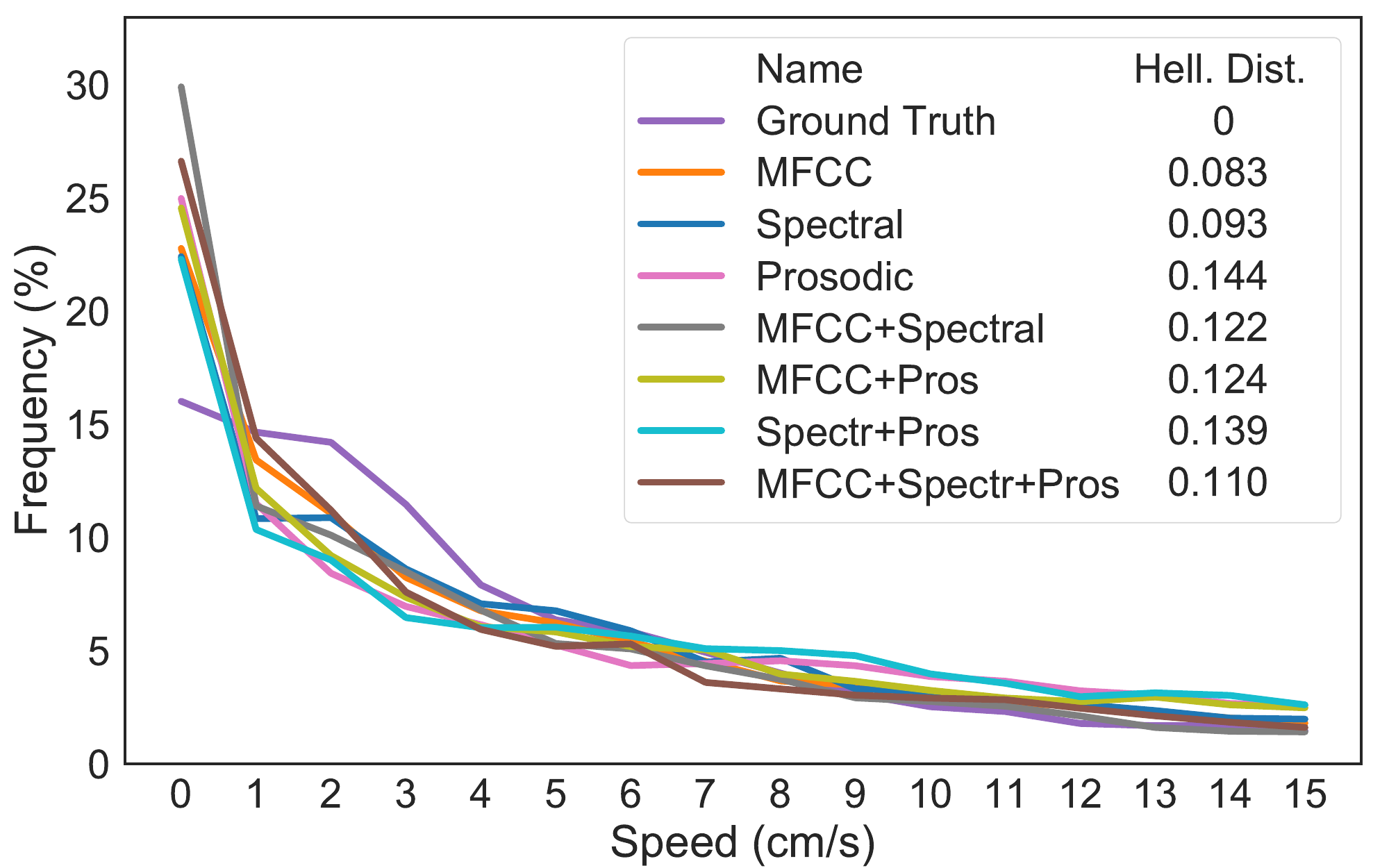}\hfill}
\subfloat[Seq.~1144: Average speed histogram.\label{sfig:avg_acc}]{\hfill\includegraphics[width=0.45\linewidth]{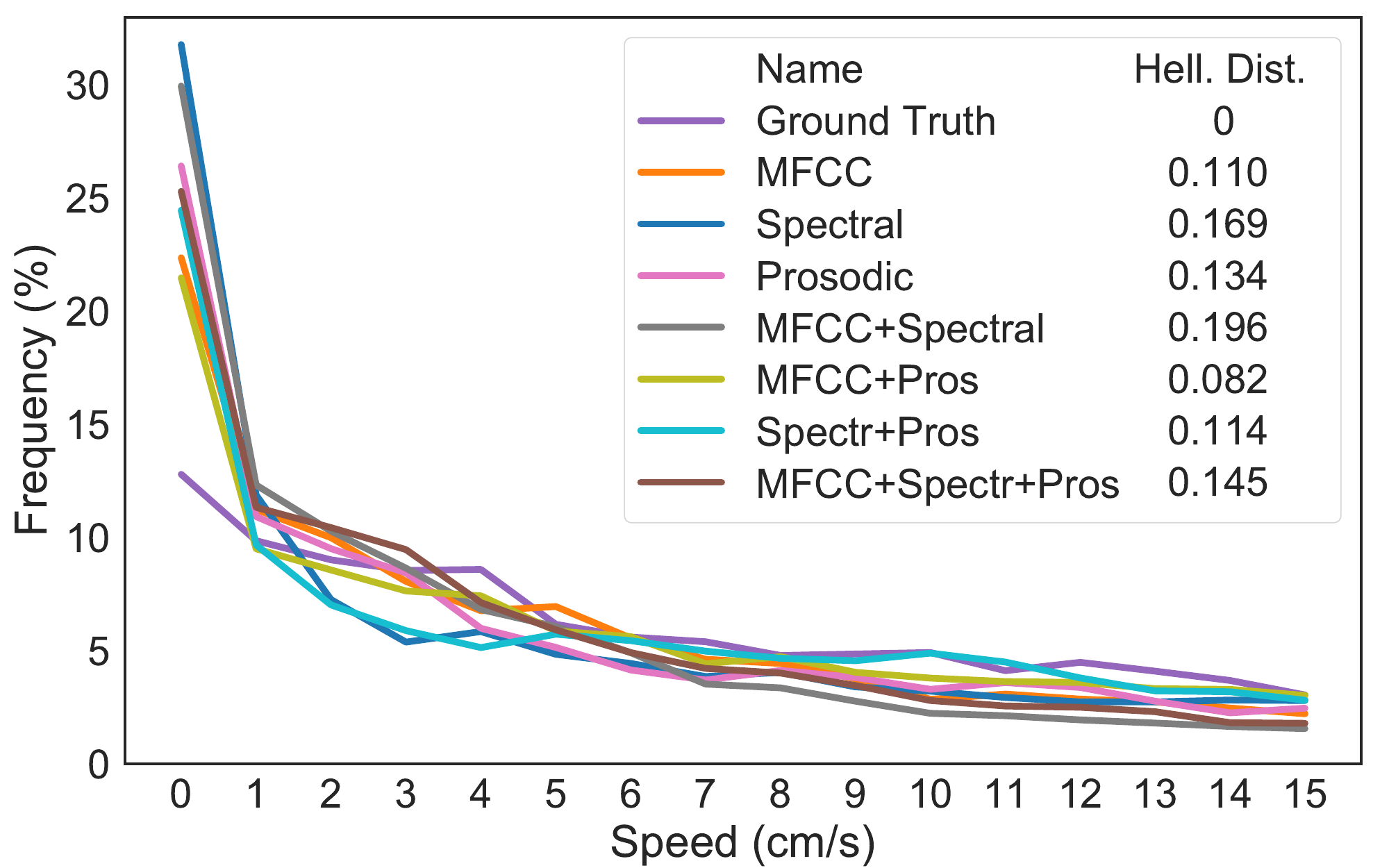}\hfill}
\\
\subfloat[Seq.~1152: Speed histogram for elbows.\label{sfig:hands_acc}]{\hfill\includegraphics[width=0.45\linewidth]{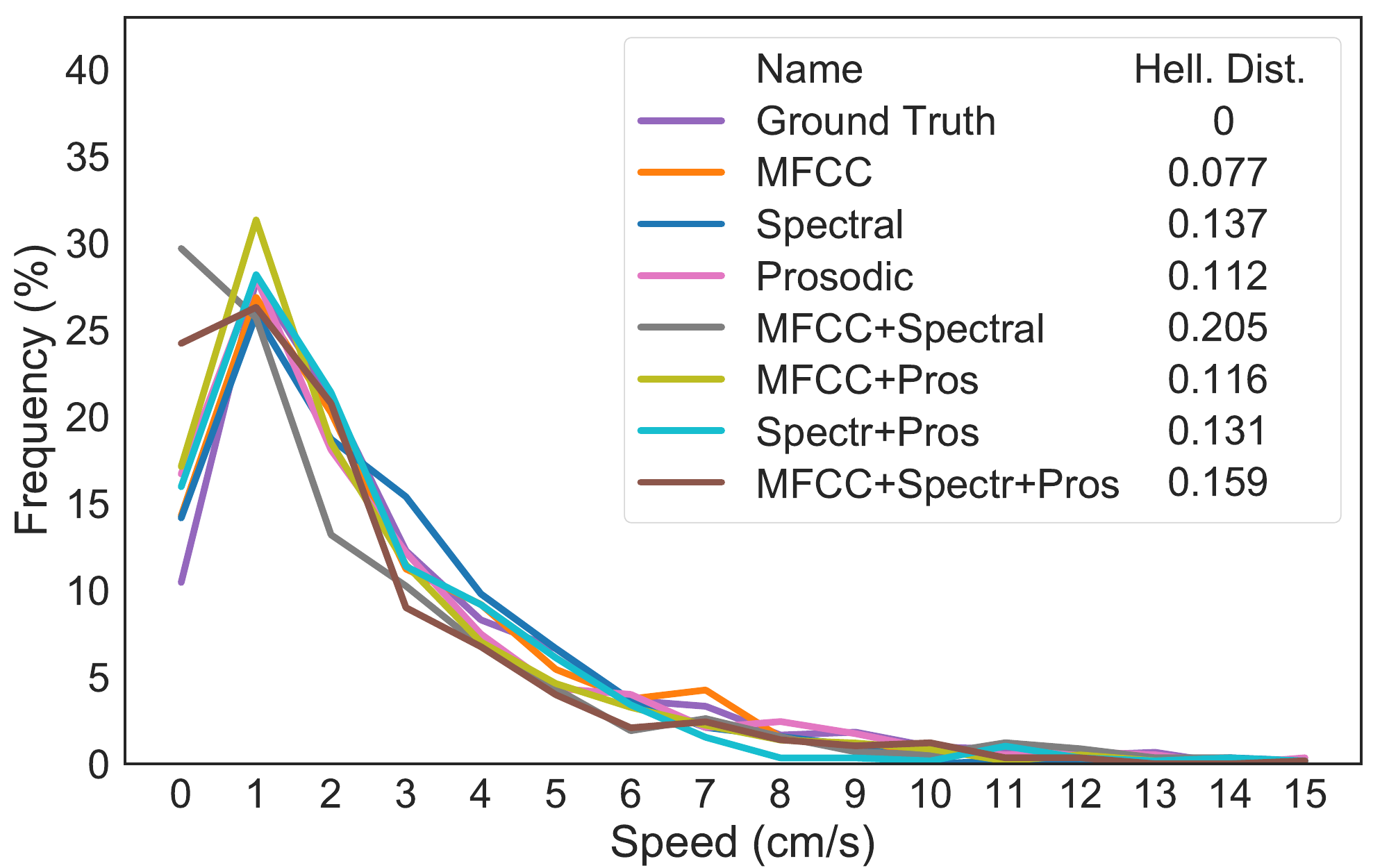}\hfill}
\subfloat[Seq.~1144: Speed histogram for elbows.\label{sfig:hands_acc}]{\hfill\includegraphics[width=0.45\linewidth]{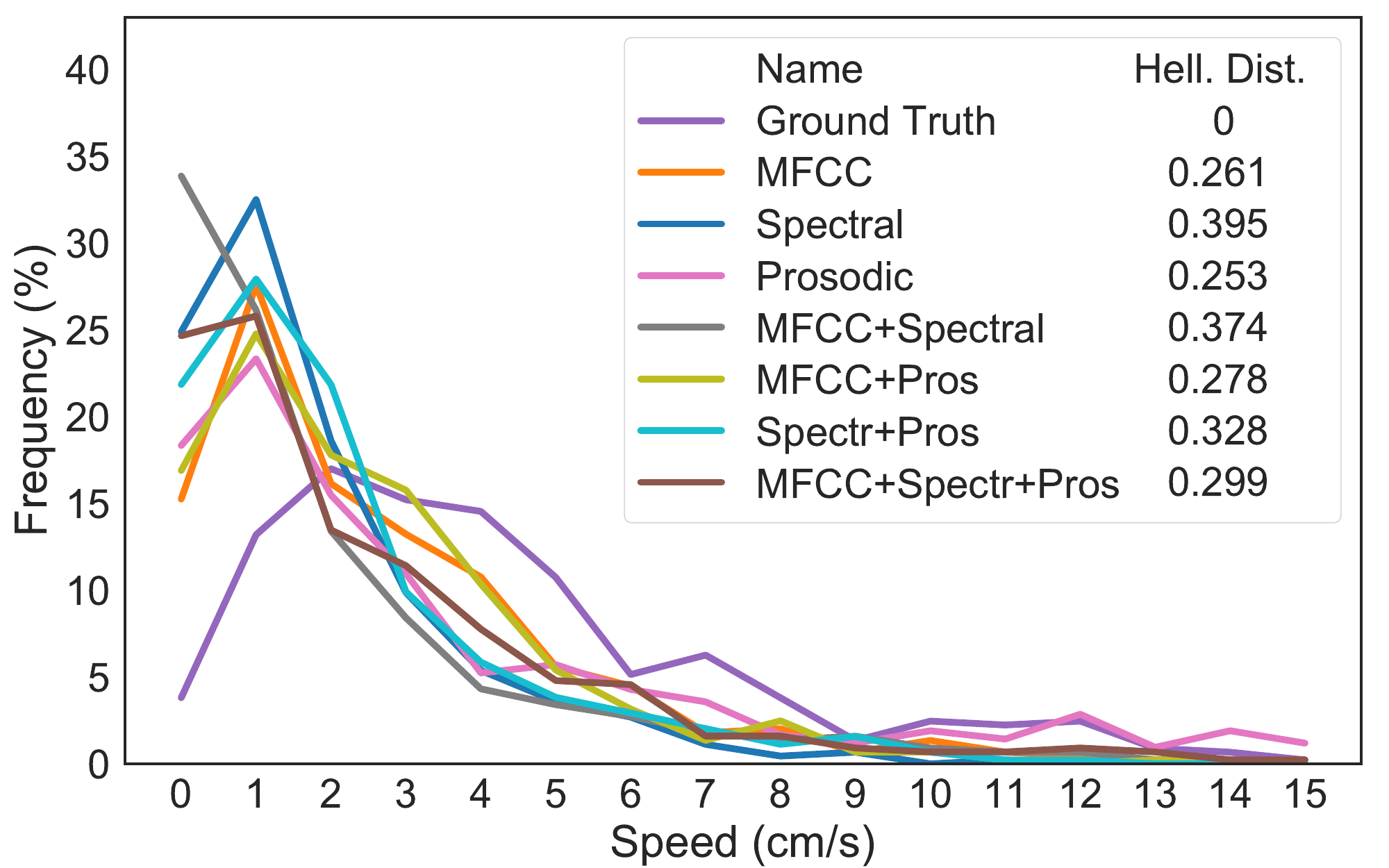}\hfill}\\
\subfloat[Seq.~1152: Speed histogram for wrists.\label{sfig:wrist_acc}]{\hfill\includegraphics[width=0.45\linewidth]{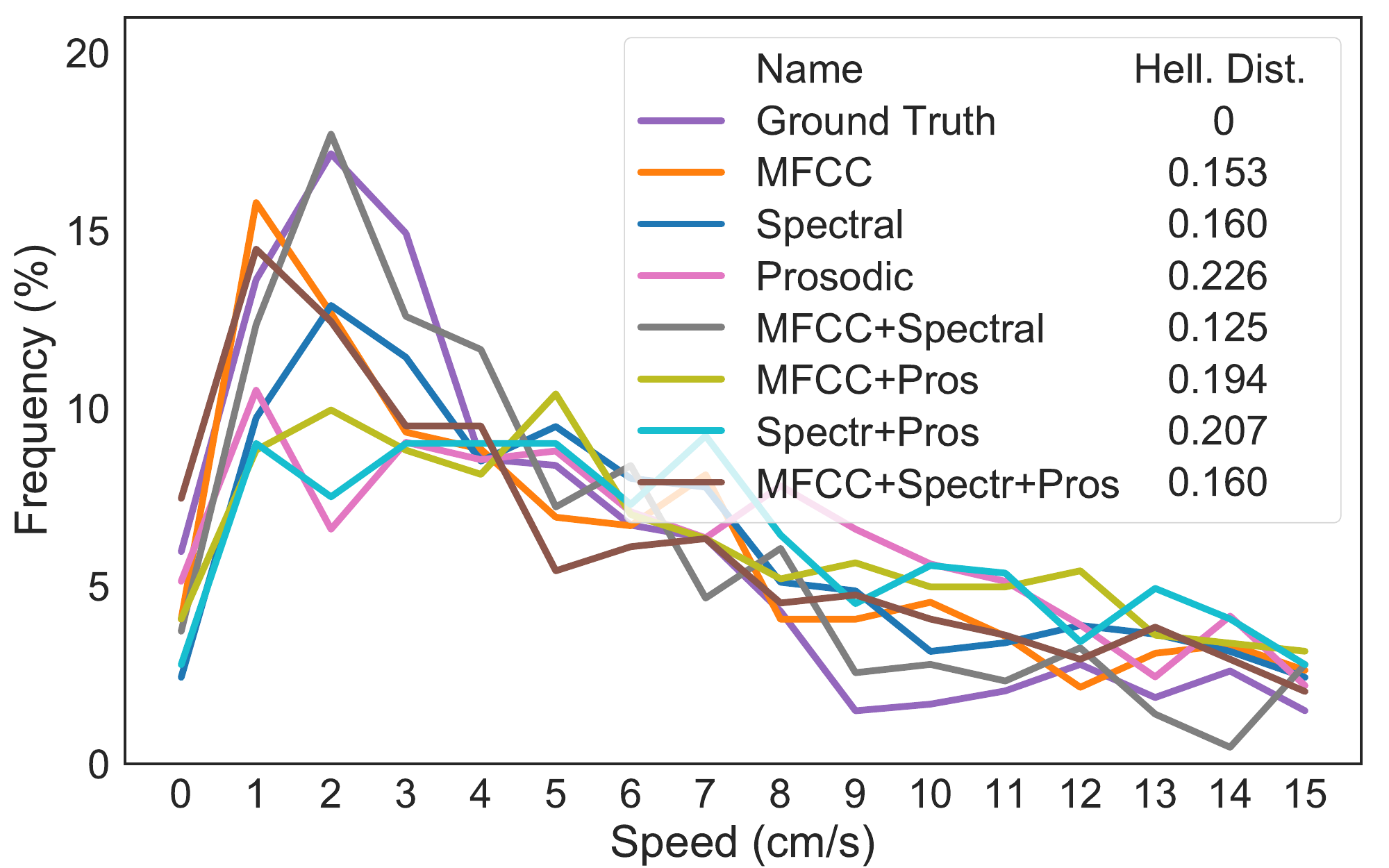}\hfill}
\subfloat[Seq.~1144: Speed histogram for wrists.\label{sfig:wrist_acc}]{\hfill\includegraphics[width=0.45\linewidth]{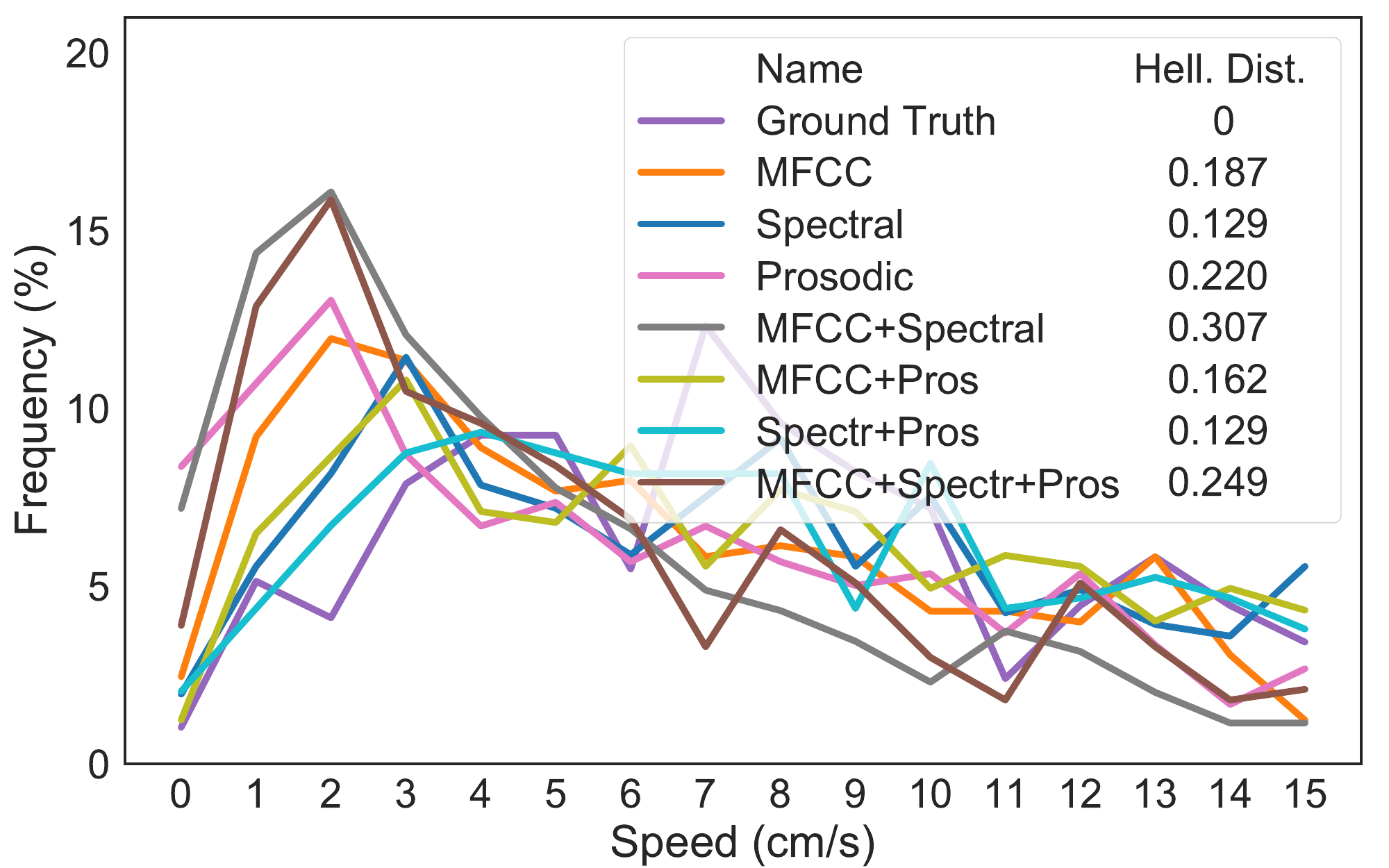}\hfill}
\\
\caption{\upd{Speed distributions for different sequences and joints.}  }
\label{fig:speed_hist_sequence_example}
\end{figure}

\upd{Table~\ref{tab:speech_sequence_features} reports average values and standard deviations of Hellinger distance between all the test sequences, considering different subsets of joints. We observe that which feature set that gives a distribution closest to natural motion differs depending on which joint on the body that is considered. We also note that the standard deviations of the values in the table are quite high, indicating substantial variability in the data. Indeed, different sequences had very different motion statistics, as demonstrated in Fig.~\ref{fig:speed_hist_sequence_example}. Histograms for all individual test sequences are available online.\footnote{Please see \href{https://doi.org/10.6084/m9.figshare.13065383}{doi.org/10.6084/m9.figshare.13065383}.}
}
\rev{Furthermore, the standard deviations for the feature sets in the table are generally larger than the differences in average Hellinger distance between the different feature sets. This suggests that differences in speed profile due to using different feature sets might not be particularly noteworthy in practice, since the variation between individual sequences when using the same feature set is substantially larger.}

\upd{To sum up, no single input-feature set performed the best for all three evaluations in Fig.~\ref{fig:speed_hist_features} and/or Table~\ref{tab:speech_sequence_features}.
\rev{There is a large amount of variability between different sequences in the data, and no clear trend across the three evaluations.}
In the absence of compelling numerical evidence in favor of choosing any particular feature set, we use MFCCs %
as input features in the remainder of this paper, since that makes our systems consistent with the baseline paper \citet{hasegawa2018evaluation}.}

\upd{As a final objective study, we compared the proposed method with the baseline on a sequence level.
The result is given in Table~\ref{tab:speech_sequence_models} and provides clear evidence that the proposed method matches the natural motion distribution better than the baseline model, although both synthetic systems exhibit notable per-sequence variation in Hellinger distance.}

\begin{table}
  \caption{\upd{Hellinger distances of speed distributions for different models, computed on a per-sequence basis. The table shows average values and standard deviations aggregated across all test sequences. For each column the values closest to the ground truth are highlighted in bold, showing that the speed distribution of the motion produced by our model is more similar to the ground truth than the other models are.}}
  \label{tab:speech_sequence_models}
  \centering
  \begin{tabular}{@{}lccc@{}}
    \toprule
    \multirow{2}{*}{Model} & \multicolumn{3}{c}{Hellinger distance w.r.t.\ ground truth} \\
     & All joints & Elbows & Wrists \\
    \toprule
    {\small Aud2Repr2Pose} & \textbf{0.133} $\pm$ 0.062 & \textbf{0.271} $\pm$ 0.147 & \textbf{0.214} $\pm$ 0.106 \\
    {\small Aud2Motion} & 0.382 $\pm$ 0.079 & 0.485 $\pm$ 0.121 & 0.372 $\pm$ 0.142 \\
    {\small Aud2Pose} & 0.325 $\pm$ 0.085 & 0.414 $\pm$ 0.091 & 0.320 $\pm$ 0.108 \\
    {\small Aud2Pose*} & 0.142 $\pm$ 0.055 & 0.440 $\pm$ 0.112 & 0.245 $\pm$ 0.088 \\
    \midrule
    {\small Ground truth} & 0 & 0 & 0 \\
    \bottomrule
\end{tabular}
\end{table}

\section{Subjective Evaluation and Discussion}
\label{sec:subj}
The most important goal in gesture generation is to produce motion patterns that are convincing to human observers.
\rev{Existing numerical measures used for evaluating gesture motion are fast and cheap to compute, but tend to be sensitive to different aspects of the motion and generally only show weak correlation with subjective performance.}
Since improvements in objective measures do not always translate into superior subjective quality for human observers, we validated our conclusions from the previous section through a number of user studies:
%
\begin{table}
  \caption{Statements evaluated in user studies. \edit{The table lists the English-language versions of the statements; the Japanese evaluation used a native Japanese version of the statements.}}
  \label{tab:eval_stat}
  \centering
  \begin{tabular}{ll}
    \toprule
    Scale & Statement \\
    \midrule
    Naturalness & Gesticulation was natural \\
      & Gesticulation was smooth \\
      & Gesticulation was fluent \\
    \midrule
    Time & Gesticulation timing was matched to speech\\
     consistency& Gesticulation speed was matched to speech \\
     & Gesticulation tempo was matched to speech\\
    \midrule
    Semantic&Gesticulation was matched to speech content\\
    consistency  & Gesticulation described the speech content well\\
      & Gesticulation helped me understand the content\\
  \bottomrule
\end{tabular}
\end{table}
two on the Japanese dataset from \citet{takeuchi2017creating} and one on the English dataset of  \citet{ferstl2018investigating}. All of them used the same questionnaire as in the baseline paper \citet{hasegawa2018evaluation}, shown in Table \ref{tab:eval_stat}. The participants watched videos and then ranked them in relation to each of the nine statements using a seven-point Likert scale from strongly disagree (1) to strongly agree (7). All models used MFCC features as input. 

\subsection{User studies on the Japanese-language dataset}
\label{ssec:user_study_j}

First
, we conducted a 1$\times$2 factorial design user study with the within-subjects factor being representation learning (Aud2Repr2Pose vs.\ Aud2Pose). We randomly selected 10 utterances from the 45 test utterances and created two videos, using each of the two gesture generation systems, for each utterance. Visual examples are provided at \href{https://vimeo.com/album/5667276}{https://vimeo.com/album/5667276}. The utterance order was fixed for every participant, but the gesture conditions were counter-balanced. 

\begin{figure}[t]
\centering
\includegraphics[width=0.9\linewidth]{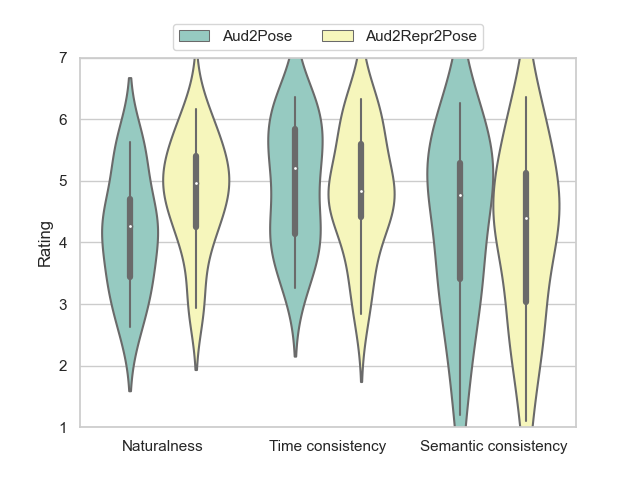}
\vspace{-1mm}
\caption{Results from the preliminary user study. We note a significant difference in naturalness only.}
\label{fig:user_study}
\end{figure}

19 native speakers of Japanese (17 male, 2 female), on average 26 years old, participated in the \edit{first} user study. A paired-sample $t$-test was conducted to evaluate the impact of the motion encoding on the perception of the produced gestures. Fig.~\ref{fig:user_study} illustrates the results we obtained for the three scales being evaluated. We found a significant difference in naturalness between the baseline (M=4.16, SD=0.93) and proposed model (M=4.79, SD=0.89), $t$=-3.6372, $p$\textless{}0.002. A 95\%-confidence interval for the mean rating improvement with the proposed system is (0.27, 1.00). There were no significant differences on the other scales: for time-consistency $t$=1.0192, $p$=0.32, for semantic consistency $t$=1.5667, $p$=0.13. \linebreak These results indicate that gestures generated by the proposed method (i.e., Aud2Repr2Pose) were perceived as more natural than the baseline (Aud2Pose).

\begin{figure}
\centering
\includegraphics[width=0.85\linewidth]{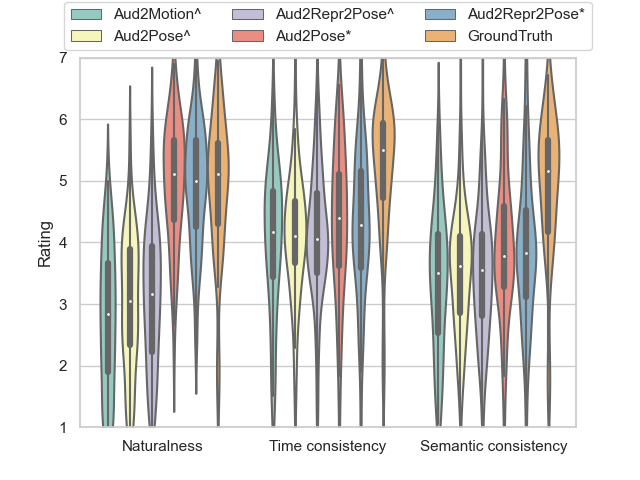}
\caption{Results of the second user study on the Japanese data: mean rating with confidence intervals (error bars). `\^{}' denotes hip-centered gestures while `*' denotes that smoothing also has been applied.}
\label{fig:japan_user_res}
\end{figure} 

After the initial user study we observed that the baseline model, Aud2Pose, exhibited many whole-body movements up and down. To remove these issues we decided to translate all motion to hip-centered coordinates and carry out a second user study to compare systems with hip-centering (denoted by a ``\^{}'' after the system name) against hip-centering and smoothing (denoted by a ``*'').
This user study used a 1$\times$6 factorial design with the within-subjects factor being the gesture generation system (Aud2Pose\^{}, Aud2Pose*, Aud2Repr2Pose\^{}, Aud2Repr2Pose*, Aud2Motion\^{} or GroundTruth). For smoothing we applied the OneEuro filter \citep{casiez20121} and sliding-window averaging with window \linebreak length 5, same as in in the baseline paper \citet{hasegawa2018evaluation}. We randomly selected 6 utterances to be tested from the 45 test utterances. The average utterance duration was 17.5 s. Examples are provided at \href{https://vimeo.com/showcase/6315181}{https://vimeo.com/showcase/6315181}. Utterance order was fixed for every participant, but the gesture conditions were randomized in a way where every condition appeared in every place in the order an equal number of times.
With 6 speech segments and 6 conditions, we obtained 36 videos. We recorded start time and end time to detect unreliable raters who did not watch the videos or listen to the audio. 

We recruited 104 native Japanese speakers through CrowdWorks. The average duration of the experiment was 28 minutes; no rater took less than 10 minutes, so none were discarded as unreliable. Participants who did not finish the questionnaire (9/104) were excluded, resulting in the data of 95 participants (57 male and 38 female) being collected. The average age of the study participants was 39.0 with a standard deviation of 10.0. Cronbach's alpha values of the questionnaire for naturalness, time consistency and semantic consistency were 0.95, 0.98 and 0.97, respectively.



We conducted a one-way ANOVA on each scale and applied Bonferroni correction to the $p$-values. The ANOVAs uncovered significant main effects on all three scales: naturalness (F(5,94)\allowbreak{}=114.4), time consistency (F(5,94)\allowbreak{}=23.0), and semantic consistency (F(5,94)\allowbreak{}=39.5).
The results of our post-hoc analyses are shown in Fig.~\ref{fig:japan_user_res}.
In terms of \textit{Naturalness} GroundTruth and all the smoothed gestures (Aud2Pos\allowbreak{}* and Aud2Repr- 2Pose*) were significantly better than all model-predicted motions without smoothing (Aud2Motion\^{}, Aud2Pose\^{} and Aud2- Repr2Pose\^{}) with $p$\textless{}0.01 for each pairwise comparison. It was not a surprise that smoothed versions were perceived as significantly more natural than non-smoothed ones, but seeing the smoothed conditions rated as natural as the ground truth motion exceeded our expectations. Apart from that, it was unexpected that there were no differences between the Aud2Pose\^{} and Aud2Repr2Pose\^{} models. The difference from the previous evaluation disappeared after we centered all the motion to the hips, removing whole-body translation. This suggests that the main problem of the Aud2Pose method in the first evaluation was that the produced gestures were jumping around. For \textit{Time-consistency} only GroundTruth was significantly better than all the other conditions ($p$\textless{}0.01). This indicates that none of the models, not even Aud2Motion\^{}, managed to model the temporal characteristics of the motion well enough. The results for \textit{Semantic-consistency} are the following: \linebreak \allowbreak Aud2Repr2Pose\allowbreak{}* \textgreater{} Aud2Motion\^{} ($p$\textless{}0.05), Aud2Pose* \textgreater{} Aud2Repr2Pose\^{} ($p$\textless{}0.05), Aud2Pose* \textgreater{} Aud2Motion\^{} ($p$\textless{}0.01), \allowbreak and GroundTruth \textgreater{} the five other conditions ($p$\textless{}0.01). Interestingly, smoothing, but not the model, is what makes the difference here, just like what we found for naturalness. We surmise that smoothness made raters appreciate the gestures more, and they had difficulties judging different aspects completely independently. \upd{This is consistent with previous findings \citep{chiu2015predicting}.} 

\subsection{User study on the English-language dataset}
\label{ssec:user_study_e}

We observed in the previous user study that smoothing plays a crucial role, so for the final user study we considered only smoothed gestures.
We conducted an experiment with a 1$\times$4 factorial design with the within-subjects factor being the model for gesture generation (Aud2Motion*, Aud2Pose*, Aud2Repr2Pose* or \linebreak GroundTruth). All  models used MFCC features as input and had their output smoothed by a moving average over 5 frames. We randomly selected 7 utterances, each roughly 10 s, from a test dataset of 20 minutes. For each utterance we created four videos, using the three gesture generation systems and the ground truth. Examples are provided at 
\href{https://vimeo.com/showcase/6287423}{https://vimeo.com/showcase/6287423}. Utterance order was fixed for every participant, but the gesture conditions were counter-balanced resulting in 24 different orders. With 7 speech segments and 4 conditions, we obtained 28 videos. We also added four additional videos: two with 
noise in the audio and two with 
noise in the video to detect unreliable raters. 
Therefore, every participant watched 32 videos.

\begin{figure}
\centering
\includegraphics[width=0.8\linewidth]{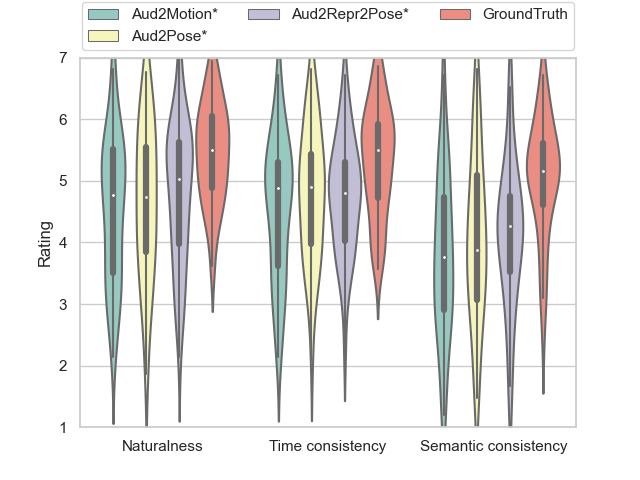}
\caption{Results of the user study on English dataset: mean rating with the confidence intervals (error bars). All the models have been smoothed.}
\label{fig:eng_user_res}
\end{figure} 

We recruited 48 native English speakers (28 male, 20 female) through Amazon Mechanical Turk, taking only those who gained a ``master'' title, completed over 10,000 HITs and had over 98\% acceptance rate. 8 participants were discarded due to not detecting the noisy videos. We replaced them by new participants. 
The average age of the participants used in the study was 44.2 with a standard deviation of 10.9.  

The result of the user study is illustrated in Fig.~\ref{fig:eng_user_res}. We conducted a one-way ANOVA on each scale and applied Bonferroni correction to the $p$-values. The ANOVAs found significant main effects in all three scales: naturalness (F(3,47)\allowbreak{}=14.4), time consistency (F(3,47)\allowbreak{}=11.4), and semantic consistency (F(3,47)\allowbreak{}=27.2).
A post-hoc analysis was conducted to evaluate the difference between each pair of conditions in each scale. On all three scales we found significant differences only between the ground truth and all the three models ($p$\textless{}0.01). 
This suggests that participants perceived all models as producing similar quality once their output was smoothed, just like we found in the Japanese-language study.
\section{Conclusions and Future Work}
\label{sec:concl}

This paper presented a new model for speech-driven gesture generation called \textit{Aud2Repr2Pose}.
Our method extends prior work on deep learning for gesture generation by applying representation learning. 
The motion representation is learned first, after which a network is trained to predict such representations from speech, instead of directly mapping speech to joint positions as in prior work. We also evaluated the effect of different representations for the 
speech.
Our code \edit{is} publicly available to encourage replication of our results.\footnote{ \edit{\href{https://github.com/GestureGeneration/Speech_driven_gesture_generation_with_autoencoder}{https://github.com/GestureGeneration/Speech\_driven\_gesture\_generation\_with\_autoencoder}}}

Our experiments show that representation learning improves the objective performance of the speech-to-gesture neural network: the proposed models match the motion distribution from the dataset better than the baseline. 
Although the proposed method was perceived as more natural than the baseline when considering the raw model output, 
appropriate post-processing (such as hip-centering and smoothing) was found to make all models in the study produce comparable quality for the study participants\upd{, across both languages considered}. This illustrates that it is non-trivial to evaluate gesture quality and gesture models, since 
model-external factors, such as smoothing, can have a dominant effect on ratings.
We believe that further research is needed to develop \edit{reliable objective} quality measures. 

Another observation from our experiments is that modifying the model to take a sequence of audio frame by frame and produce a sequence of gestures frame by frame, Aud2Motion, did not result in smooth motion. 

\subsection{Limitations}
The main limitation of our approach, as with any data-driven method and particularly those based on deep learning, is that it requires substantial amounts of parallel speech-and-motion training data of sufficient quality in order to \edit{achieve} good prediction performance. Apart from that, the learned gestures will depend greatly on the contents of the data: if the actor in the dataset produces a small range of gestures, the model is likely to also generate a small range of gestures; if the actor is producing only beat gestures, the model can only learn such gestures.
\upd{This includes both individual differences, but also other important variation such as cultural differences. Essentially, the model is restricted to learn the gesture style of the actor(s) in the training database.}
In the future, we might overcome this limitation by obtaining very large \upd{multi-speaker} datasets directly from publicly-available video recordings using motion-estimation techniques as in \citet{yoon2018robots,jonell2019learning,yoon2020speech}\upd{, and by learning to embed different speakers and gesture styles in a space, e.g., following recent work by \citet{yoon2020speech}}.

\rev{There were also limitations in our evaluation. Most prominently, different evaluation metrics seemed to display a preference for different speech features, and no feature set performed better than the others across the board. We also note that, to date, there are no good objective metrics of the quality of gesture-generation models and the most commonly-used metrics do not align well with perceptual quality; see, for instance, \cite{kucherenko2021large}. Taken together, these two observations mean that we cannot draw any strong conclusions regarding which speech feature set is the best for the Aud2Repr2Pose gesture-generation model.}

\subsection{Future work}
\label{ssec:future_work}

We see several 
directions for future research:

    
Firstly, text should be taken into account, e.g., as in \citet{kucherenko2020gesticulator}.
Gestures that co-occur with speech depend greatly on the semantic content of the utterance. Our model generates mostly beat gestures, as we rely only on speech acoustics as input. Hence the model can benefit from incorporating the text transcription of the utterance along with the speech audio, especially since semantically-informative representations can be obtained from models pre-trained on large amounts of text alone. This may enable producing a wider range of gestures (also metaphoric, iconic and deictic gestures). 
    
Secondly, the learned model can be applied to improve human-robot interaction by enabling robots (for instance the NAO robot) to accompany their speech by appropriate co-speech gestures, like in \citet{yoon2018robots}.

\edit{Thirdly, the model could be made probabilistic to enabling randomly sampling different gestures for the same speech input, like in \citet{alexanderson2020style}.}

\upd{Finally, it would be interesting to explore sequence representations for MotionE and MotionD\rev{, similar to the recent work of \citet{jinhong_lu_2020_4088376}}, instead of representing just one frame and its derivative.} \rev{More generally, the temporal character of the model with respect to the representation learning could also be investigated.}

\section*{Acknowledgements}
The authors would like to thank Jan Gorish for review and feedback on the manuscript and Sanne van Waveren, Patrik Jonell and Iolanda Leite for insightful discussions. \upd{We are also grateful to the anonymous reviewers for their helpful suggestions.}


\section*{Funding}
This work was partially supported by the Swedish Foundation for Strategic Research Grant No.\ RIT15-0107 (EACare), by the Wallenberg AI, Autonomous Systems and Software Program (WASP) funded by the Knut and Alice Wallenberg Foundation, and by JSPS Grant-in-Aid for Young Scientists (B) Grant Number 17K18075.

\balance
\bibliographystyle{apacite}
\bibliography{bibliography}

\pagebreak

\begin{wrapfigure}[8]{l}{25mm} 
    \includegraphics[width=1in,height=1.25in,clip,keepaspectratio]{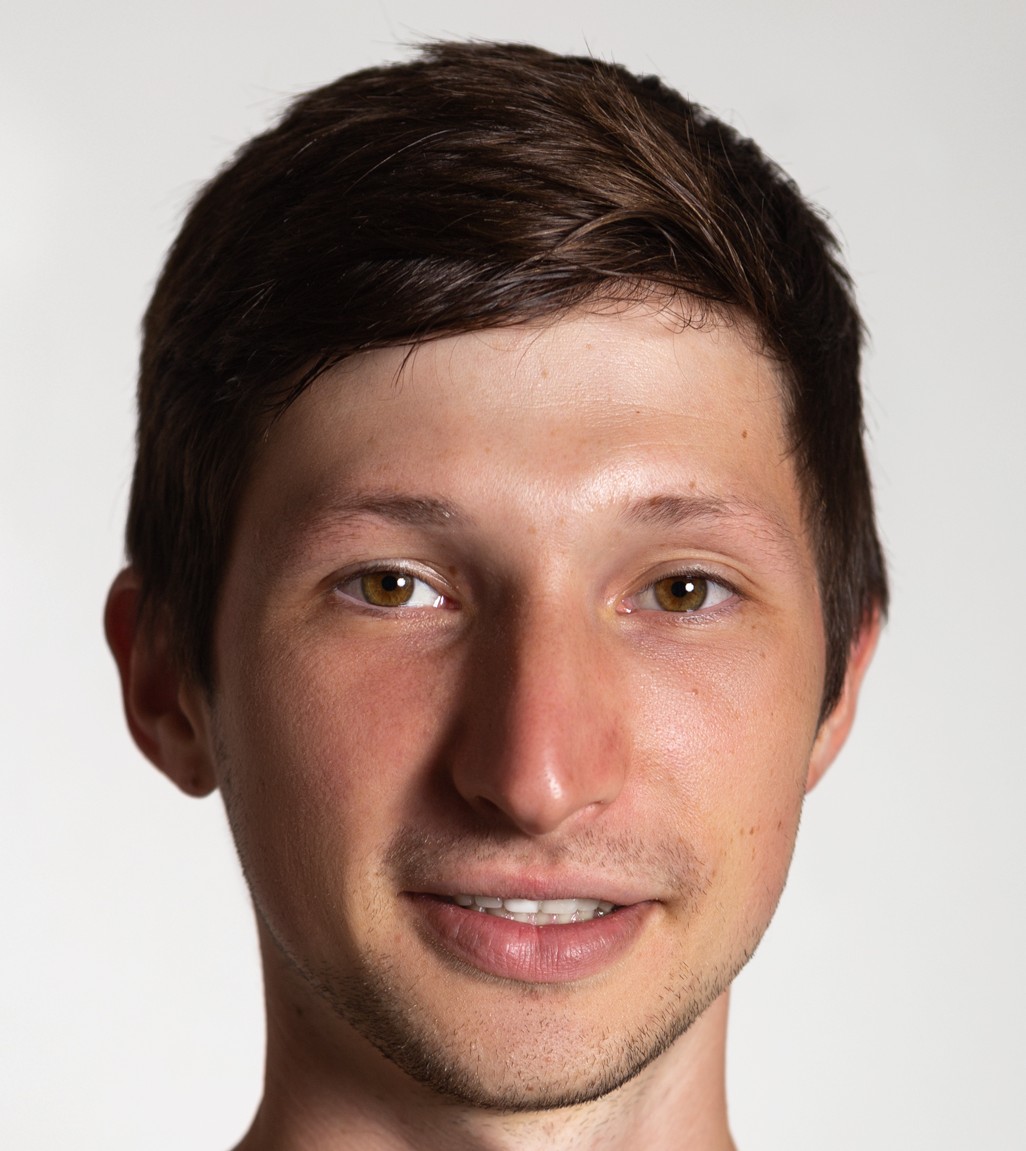}
  \end{wrapfigure}\par
  \noindent \upd{\textbf{Taras Kucherenko} is a Ph.D.~student in the Division of Robotics, Perception and Learning (RPL) at KTH Royal Institute of Technology in Stockholm, Sweden. He received his M.Sc.\ in Machine Learning at RWTH Aachen with an emphasis on Natural Language Processing. His B.Sc\ was in applied math at KPI, Kyiv, with an emphasis on cryptography. His current research is about building generative models for non-verbal behavior generation to enable social agents to use body language as an additional communication tool.} \par

\vspace{12mm}
\begin{wrapfigure}[7]{l}{25mm} 
\vspace*{-\intextsep}
    \includegraphics[trim=200 200 200 0,width=1in,height=1.25in,clip,keepaspectratio]{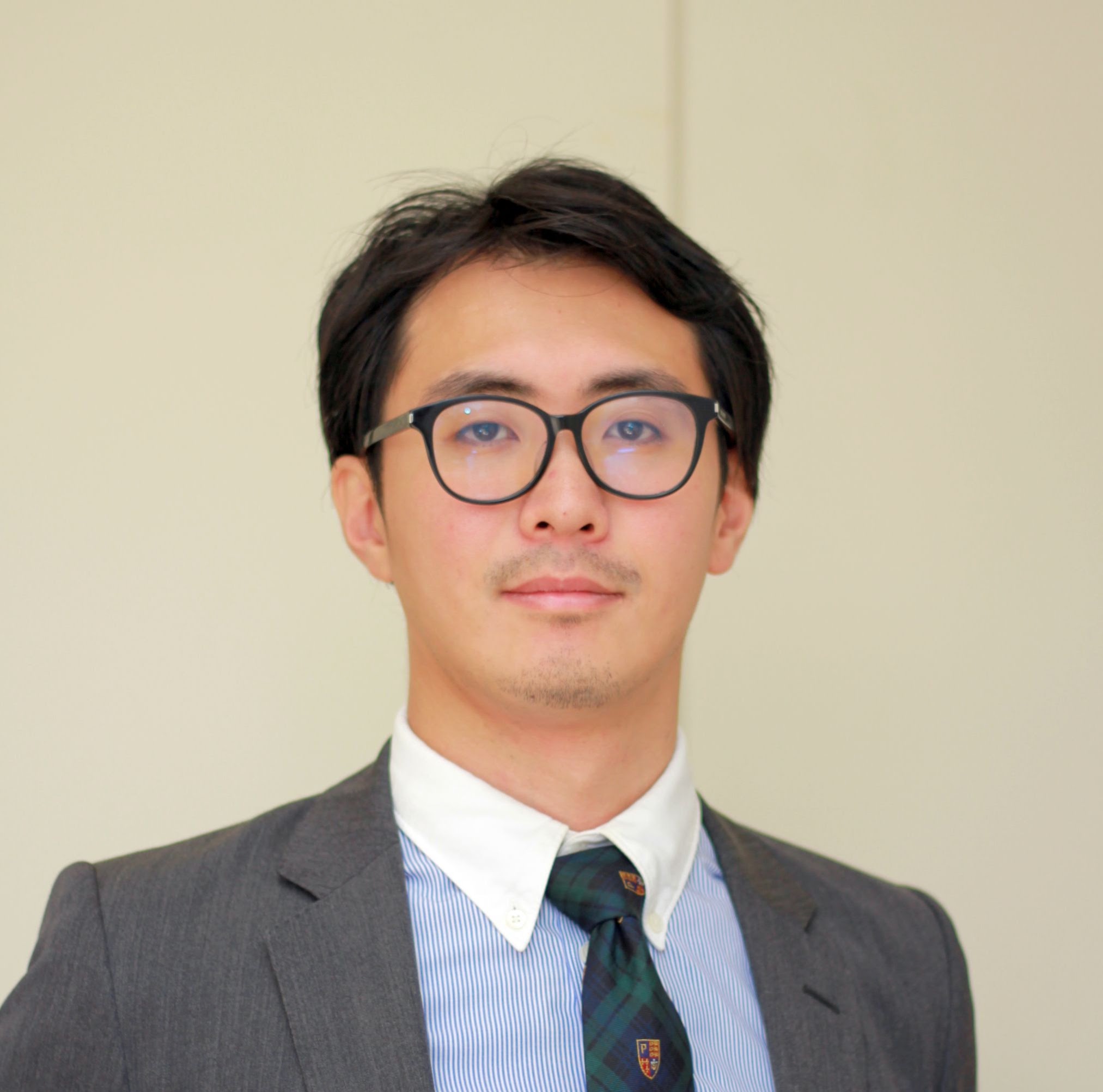}
  \end{wrapfigure}\par
  \noindent \upd{\textbf{Dai Hasegawa} is an Associate Professor at the Faculty of Engineering, Hokkai Gakuen University, Sapporo, Japan. He received his Ph.D.\ in Information Science at Graduate School of Information Science and Technology, Hokkaido University. He is interested in building embodied conversational agents/social robots and their application in education. Currently, he is working on non-verbal behavior generation using data-driven approaches.} \par

\vspace{12mm}
\begin{wrapfigure}[8]{l}{25mm} 
\vspace*{-\intextsep}
    \includegraphics[trim=50 0 50 0,width=1in,height=1.25in,clip,keepaspectratio]{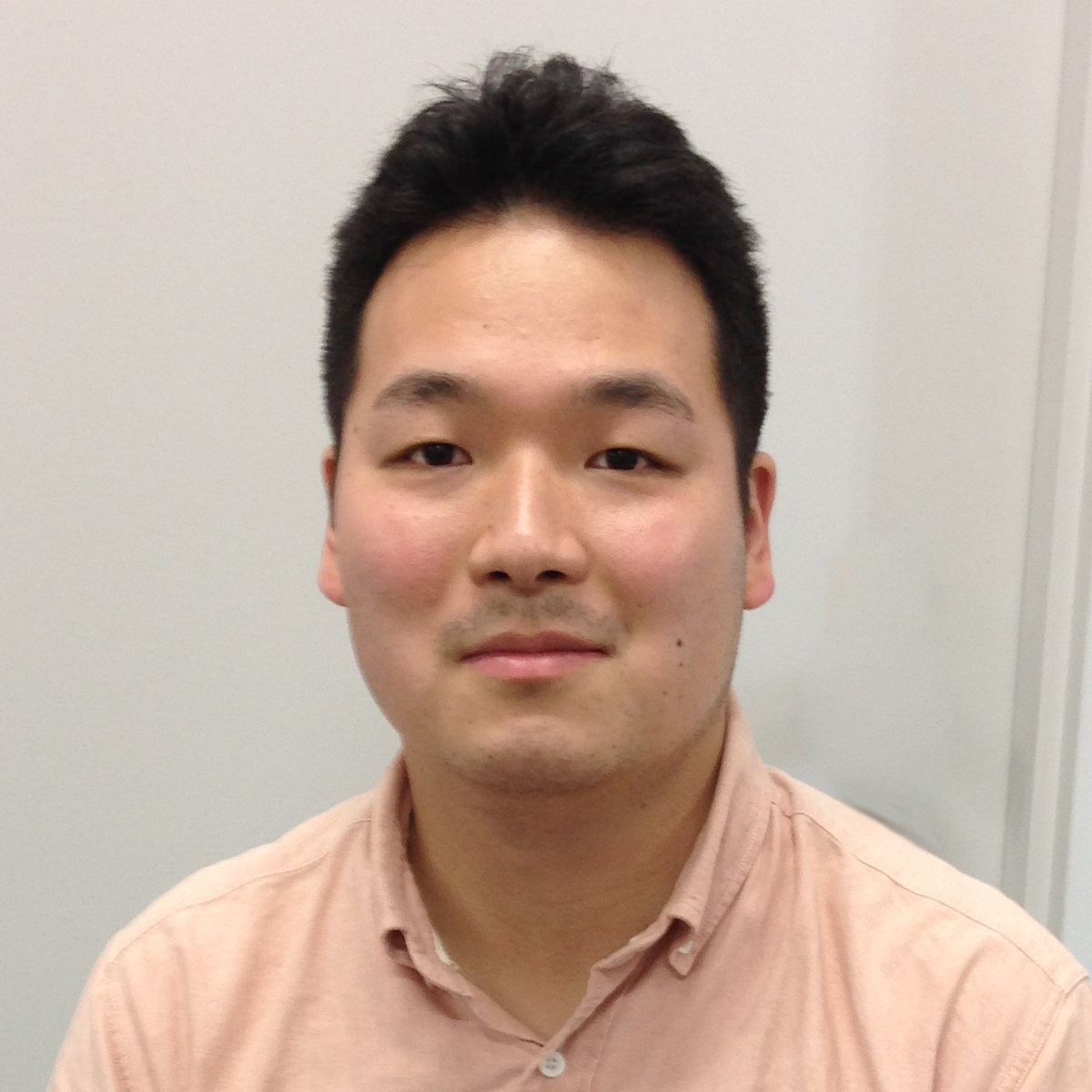}
  \end{wrapfigure}\par
  \noindent \upd{\textbf{Naoshi Kaneko} is an Assistant Professor at the Department of Integrated Technology, Aoyama Gakuin University, Sagamihara, Japan. He holds an M.Sc.\ and a Ph.D.\ in Engineering, both from the Intelligent and Information Course at Graduate School of Science and Engineering, Aoyama Gakuin University. His research interests lie in computer vision and machine learning, with a current emphasis on estimating and generating human pose, body shape, and non-verbal behaviors.} \par
\vspace{12mm}
\begin{wrapfigure}[7]{l}{25mm} 
\vspace*{-\intextsep}
    \includegraphics[trim=0 21 0 21,width=1in,height=1.25in,clip,keepaspectratio]{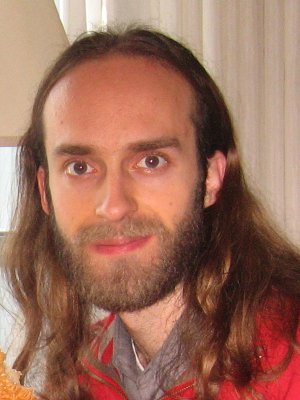}
  \end{wrapfigure}\par
  \noindent \upd{\textbf{Gustav Eje Henter} is an Assistant Professor in Intelligent Systems with specialization in Machine Learning in the Division of Speech, Music and Hearing (TMH) at KTH Royal Institute of Technology in Stockholm, Sweden. He received his Ph.D.\ in Electrical Engineering from KTH, followed by post-doc positions at the University of Edinburgh in Edinburgh, Scotland, and at the National Institute of Informatics in Tokyo, Japan, before returning to KTH in 2018. His research interests focus on probabilistic modeling and machine learning for data-generation tasks, especially motion and gesture generation and statistical speech synthesis, as well as how such approaches can be evaluated.} \par

\vspace{12mm}
\begin{wrapfigure}[6]{l}{25mm} 
\vspace*{-\intextsep}
    \includegraphics[trim=0 21 0 21,width=1in,height=1.25in,clip,keepaspectratio]{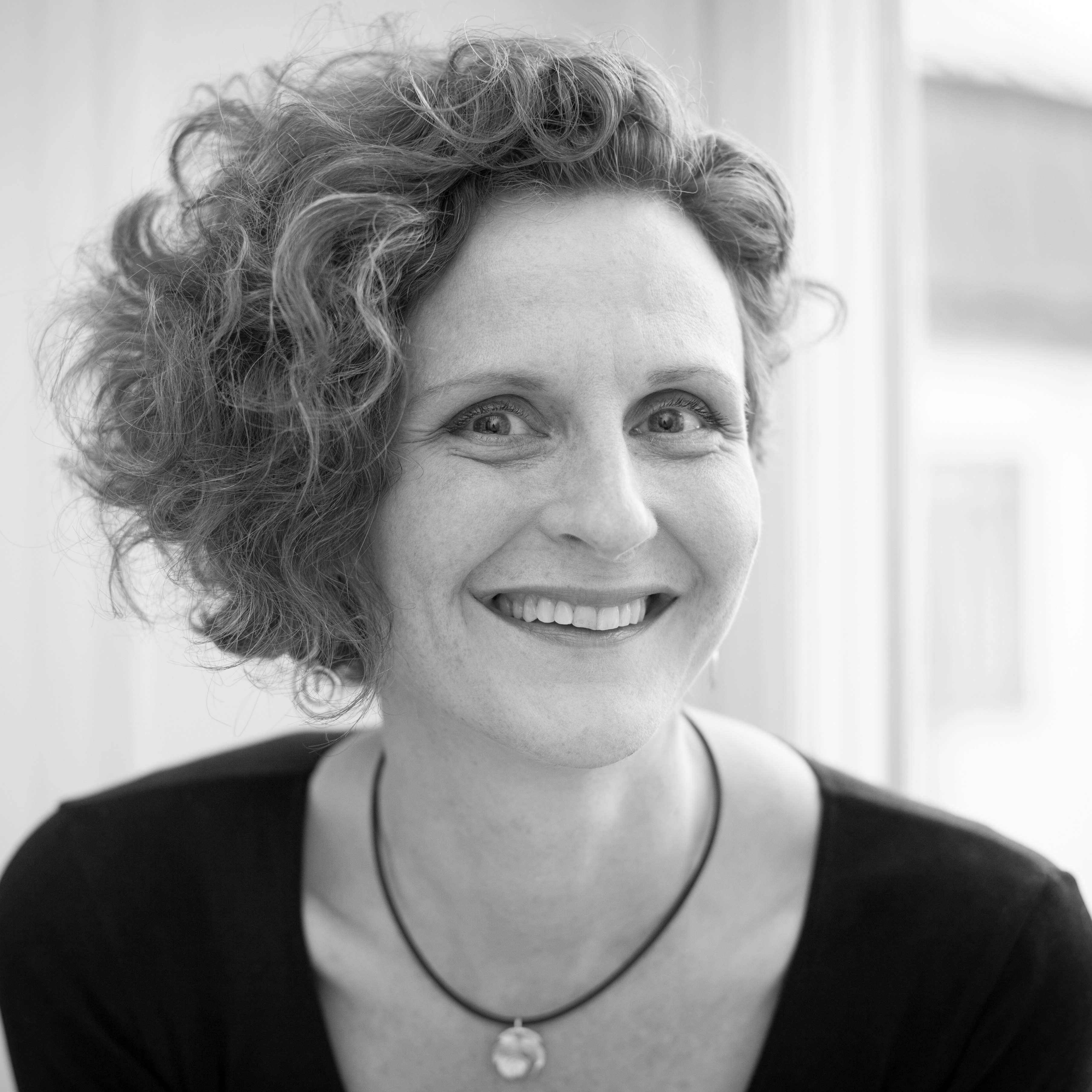}
  \end{wrapfigure}\par
  \noindent \upd{\textbf{Hedvig Kjellstr{\"o}m} is a Professor of Computer Science in the Division of Robotics, Perception and Learning at KTH Royal Institute of Technology in Stockholm, Sweden. She received her Ph.D.~from KTH in 2001.
  After this she worked as a scientist at the Swedish Defence Research Agency, returning to KTH in 2007. Her present research focuses on methods for enabling artificial agents to interpret the behavior of humans and other animals, and also to behave in ways interpretable to humans. She is recipient of the Koenderink Prize for fundamental contributions to Computer Vision 2010.} \par

\end{document}